\documentclass[11pt]{article}

\usepackage[margin=1in]{geometry}

\usepackage{marginnote} 

\usepackage{xcolor}
\definecolor{navy}{rgb}{0, 0, 0.75}

\usepackage{graphicx}
\usepackage{tikz}

\usepackage{caption}
\usepackage{subcaption}
\usepackage{longtable}

\usepackage{amsmath, amsthm, amssymb}
\usepackage{bm}

\usepackage[shortlabels]{enumitem}

\usepackage{algorithm}
\usepackage[indLines=false]{algpseudocodex}

\usepackage[colorlinks=true,linktoc=page,allcolors=navy]{hyperref}
\usepackage{zref-clever} 
\zcsetup{
    cap=true,
    nameinlink=false
}
\newcommand{\cref}{\zcref}
\newcommand{\Cref}{\zcref}

\usepackage[overload]{keytheorems}

\theoremstyle{plain}
\newtheorem{theorem}{Theorem}
\numberwithin{theorem}{section}
\newtheorem{corollary}[theorem]{Corollary}
\newtheorem{lemma}[theorem]{Lemma}
\newtheorem{claim}[theorem]{Claim}

\theoremstyle{definition}
\newtheorem{definition}[theorem]{Definition}
\newtheorem{remark}[theorem]{Remark}
\newtheorem{example}[theorem]{Example}

\DeclareMathOperator*{\E}{\mathbb{E}}
\DeclareMathOperator*{\Var}{\mathrm{Var}}
\DeclareMathOperator*{\argmin}{argmin}

\newcommand{\Beta}{\mathrm{Beta}}
\newcommand{\Unif}{\mathrm{Uniform}}

\newcommand{\proj}{\mathrm{proj}}

\newcommand{\eps}{\varepsilon}
\renewcommand{\epsilon}{\varepsilon}
\renewcommand{\phi}{\varphi}
\newcommand{\eat}[1]{}
\newcommand{\rn}{\bm} 

\newcommand{\A}{\mathcal{A}}
\newcommand{\C}{\mathcal{C}}
\newcommand{\D}{\mathcal{D}}

\renewcommand{\L}{\mathcal{L}}

\newcommand{\N}{\mathbb{N}}
\newcommand{\R}{\mathbb{R}}
\newcommand{\V}{\mathcal{V}}
\newcommand{\W}{\mathcal{W}}
\newcommand{\X}{\mathcal{X}}

\newcommand{\ECE}{\mathrm{ECE}}
\newcommand{\TwoCE}{\mathrm{CE}_2}
\newcommand{\CE}{\mathrm{CE}}

\newcommand{\smCE}{\mathrm{smCE}}
\newcommand{\x}{\rn{x}}
\newcommand{\y}{\rn{y}}
\newcommand{\z}{\rn{z}}

\renewcommand{\le}{\leqslant}
\renewcommand{\ge}{\geqslant}

\newcommand{\poly}{\mathrm{poly}}

\newcommand{\abs}[1]{\lvert#1\rvert}

\newcommand{\sign}{\mathrm{sign}}

\usepackage{accents}

\usepackage{cancel}
\usepackage{nicefrac}

\usepackage{comment}

\newif\ifdraft
\draftfalse
\newcommand{\mnote}[3]{%
    \ifdraft%
  {\color{#1}{\textsuperscript{*}}}%
  \marginnote{\raggedright\footnotesize\color{#1}{#2: #3}}%
  \fi%
}
\newcommand{\vkmnote}[1]{\mnote{blue!70!black}{VK}{#1}}%
\newcommand{\pgmnote}[1]{\mnote{blue!70!black}{PG}{#1}}%
\newcommand{\scmnote}[1]{\mnote{blue!70!black}{SC}{#1}}%

\newcommand{\EE}{\scalebox{1.2}{$\mathbb{E}$}}

\renewcommand{\P}{\mathcal{P}}
\newcommand{\Q}{\mathcal{Q}}

\algrenewcommand\algorithmicdo{\!\!:}
\algrenewcommand\algorithmicthen{\!\!:}
\algrenewcommand\algorithmicelse{\textbf{else}:}
\algrenewtext{Function}[2]{\textbf{def} #1(#2):}

\title{Agnostic Learning by Imitating}
\author{
    S\'ilvia Casacuberta\thanks{Supported by a Knight-Hennessy Scholarship, the Simons Foundation Collaboration on the Theory of Algorithmic Fairness, and the Simons Foundation investigators award 17351.} \\
    Stanford University 
\and
    Parikshit Gopalan \\
    Apple
\and
    Varun Kanade\thanks{This research has been supported in part by EPSRC grant EP/Y028872/1, Mathematical Foundations of Intelligence: An ``Erlangen Programme'' for AI.} \\
    Oxford University
\and
    Omer Reingold\thanks{Supported by the Simons Foundation Collaboration on the Theory of Algorithmic Fairness, and the Simons Foundation investigators award 17351.} \\
    Stanford University
\and
    Konstantinos Stavropoulos\thanks{This work was done in part while the author was an intern at Apple and while at The University of Texas at Austin. The author was supported in part by the NSF AI Institute for Foundations of Machine Learning (IFML) and, during the 2026–27 academic year, by the Institute for Advanced Study and an anonymous donor.} \\
    Institute for Advanced Study
\and  
    Pranay Tankala\thanks{Work done during an internship at Apple, as well as with support from Simons Foundation Grant 733782 and Cooperative Agreement CB20ADR0160001 with the United States Census Bureau.} \\
    Harvard University
}
\date{November 3, 2026}

\begin{document}

\maketitle

\thispagestyle{empty}

\begin{abstract}
    
    Given a class of machine learning models and sample access to a data distribution with binary labels, which computational task is harder: training a discriminative model that distinguishes positive and negative labels as well as the best member of the class, or training a generative model whose outputs are indistinguishable from real data by all members of the class?
    We prove that these two fundamental tasks in computational learning theory are not just closely related, but actually equivalent to each other in complexity. One direction of this equivalence was previously known. Indeed, the powerful \emph{regularity lemma} in complexity theory gives an efficient oracle reduction from the indistinguishability task (a.k.a. \emph{multiaccuracy}) to the distinguishing task (a.k.a. \emph{agnostic learning}) in the style of boosting algorithms. Our main contribution is an efficient oracle reduction in the other direction.
    
    Taking this idea further, we show that a multiaccuracy oracle is a remarkably powerful primitive, achieving not just agnostic learning but also a wide range of seemingly more demanding learning objectives via similarly efficient and direct constructions. Our results crucially use the power of making adaptive queries to the oracle, and stand in stark contrast to prior work, which showed that post-processing a single multiaccurate predictor is insufficient for even weak forms of learning. On a technical level, we draw a connection between our (offline) problem and a version of no-regret learning in the online setting, which allows us to use follow-the-leader type algorithms to modify the label distribution and aggregate the predictions of the multiaccuracy oracle. 
\end{abstract}

\newpage
\thispagestyle{empty}
\tableofcontents
\newpage
\pagestyle{plain}
\setcounter{page}{1}

\section{Introduction}

Suppose we are given a class $\C$ of machine learning models. These can either be viewed abstractly as an arbitrary set of functions, or more concretely as a set of decision trees (depth $\le d$), neural networks (parameters $\le p$), or Boolean circuits (size $\le s$), to name a few examples. Suppose also that we have been given sample access to a ground-truth distribution $\D$ over labeled data points $(\rn{x}, \rn{y})$, where $\rn{x} \in \X$ is a feature vector and $\rn{y} \in \{\pm 1\}$ is a binary label. In this setting, there are two fundamental and natural computational goals that one might attempt to achieve:
\begin{enumerate}
	\item Train a discriminative model that \emph{distinguishes} positively and negatively labeled data points from $\D$ as well as (or better than) the best model in the baseline class $\C$.
	\item Train a generative model which produces synthetic labels for each point, such that the resulting joint distribution of points and labels is \emph{indistinguishable} from real labeled data drawn from $\D$ by all observers in the class $\C$.
\end{enumerate}

Although these two tasks sound related, they are not quite the same, and they lie at the core of two different formulations of the question: \emph{What does it mean to learn optimally, relative to a baseline class?} The first task, namely the distinguishing task with respect to a restricted class of competitors, corresponds to the agnostic model for learning, introduced by \cite{haussler1992decision} and \cite{kearns1994toward}, extends Valiant's influential theory of Probably Approximately Correct (PAC) learning \cite{valiant1984pac} beyond the realizable setting. The second task, which frames learning as ensuring indistinguishability with respect to a restricted class of distinguishers, has its roots in the imitation game proposed by Alan Turing as a test of machine intelligence, now known as the Turing test \cite{turing1950test}. The form of indistinguishability that is relevant to us is multiaccuracy, which was first defined in the context of complexity theory \cite{TrevisanTV09}, and later in the setting of algorithmic fairness \cite{hkrr2018}.

In this work, we ask: What is the complexity of these two learning tasks relative to one another? That is, which of the two tasks is easier and which is harder in terms of its computational complexity? 
Also, does the answer change depending on how restricted or how expressive the class $\C$ is, or depending on the properties of the underlying data distribution $\D$?

Before stating our results, let us briefly state the formal definitions of the versions of the distinguishing and indistinguishability tasks that we consider.

\paragraph{Distinguishing.} 
We model the distinguishing task as \emph{agnostic learning} \cite{haussler1992decision,kearns1994toward}. Here, we are given independent samples from $\D$ and a description of a class $\C$ of functions $c : \X \to [-1, +1]$, and we train a discriminative \emph{model} or \emph{predictor} $p : \X \to [-1, +1]$. Each output of $p$ is a number between $-1$ and $+1$, which should be interpreted as the mean of a distribution over the two-element set of binary labels $\{-1, +1\}$. Our goal is for the expected absolute value of the error $\rn{y} - p(\rn{x})$ to be not much worse than that of the best model in the class $\C$:
\begin{equation}
\label{eq:agnostic-learning-intro}
    \E\bigl[|\rn{y} - p(\rn{x})|\bigr] \le \min_{c \in \C}\, \E\bigl[|\rn{y} - c(\rn{x})|\bigr] + \eps.
\end{equation}
Informally, our model should ``beat'' the baseline class $\C$ at loss minimization. Our algorithm should succeed with probability at least $1 - \delta$ regardless of the conditional distribution of $\rn{y}$ given $\rn{x}$, but we treat the marginal distribution $\D_\X$ over the features $\rn{x}$ as fixed and known in advance.

Although we have stated the definition of agnostic learning in terms of $\ell_1$ loss minimization, it can be stated just as easily in terms of distinguishing two distributions over the feature space $\X$, corresponding to positively and negatively labeled examples. The first distribution is the conditional distribution $\D_+$ of $\rn{x}$ given $\rn{y}=+1$ and the second is the conditional distribution $\D_-$ of $\rn{x}$ given $\rn{y} = -1$. If labels are balanced, meaning $\E[\rn{y}]=0$, then a model $p$ satisfies inequality \eqref{eq:agnostic-learning-intro} if and only if its advantage at distinguishing $\D_+$ from $\D_-$ is at most $2\eps$ worse than the best distinguishing advantage of any function in $\C$. Here, distinguishing advantage has the usual definition of
\[
    \mathrm{Adv}_f(\P, \Q) = \EE_{\rn{x} \sim \P} \bigl[f(\rn{x})\bigr] - \EE_{\rn{x} \sim \Q}\bigl[f(\rn{x})\bigr],
\]
for probability distributions $\P$ and $\Q$ on $\X$ and a test $f: \X \rightarrow [-1, +1]$.

\paragraph{Indistinguishability.}
We model the indistinguishability task as \emph{multiaccuracy} \cite{TrevisanTV09,hkrr2018}. Here, $\D$ and $\C$ are the same as in agnostic learning, but we now view $p : \X \to [-1, +1]$ as a generative model, a perspective that we will soon elaborate. Our goal is for the error, still $\rn{y} - p(\rn{x})$, to be approximately uncorrelated with all models in the class $\C$:
\begin{equation}
\label{eq:multiaccuracy-intro}
    \max_{c \in \C}\,\Bigl\lvert \E\bigl[c(\rn{x})(\rn{y} - p(\rn{x}))\bigr] \Bigr\rvert \le \eps.
\end{equation}
Informally, our model should ``fool'' observers in the class $\C$ into thinking that the labels it generates are the same as real labels. Once again, we allow a small failure probability $\delta$, treat the conditional distribution of $\rn{y}$ given $\rn{x}$ as unknown, and treat the marginal distribution of $\rn{x}$ as known. 

Although we have stated the definition of multiaccuracy in terms of correlation, it can be stated just as easily in terms of distinguishing the ground-truth distribution $\D$ over real data points $(\rn{x}, \rn{y})$ from the modeled distribution $\tilde{\D}$ over simulated data points $(\rn{x}, \tilde{\rn{y}})$, where $\tilde{\rn{y}}$ is generated from $p$ in the sense that $\E[\tilde{\rn{y}}|\rn{x}] = p(\rn{x})$. Indeed, a two-input distinguisher $A(x, y)$ taking values in $[-1, +1]$ achieves absolute advantage at most $\eps$ at distinguishing $\D$ from $\tilde{\D}$ if and only if $p$ satisfies inequality \eqref{eq:multiaccuracy-intro} with respect to the function $c_A(x) = (A(x, +1) - A(x, -1))/2$.

The {\em imitation} in the title refers to multiaccuracy, since the only guarantee we have is that the real and simulated distributions look the same to observers modeled by functions in $\C$.

\subsection{Main Result}

Our main result is that distinguishing (agnostic learning) and indistinguishability (multiaccuracy) are statistically and computationally equivalent in complexity for all feature distributions $\D_\X$ and all model classes $\C$. By ``equivalent'', we mean that an efficient algorithm for either one of the two problems can be transformed into an efficient algorithm for the other using a simple and elegant oracle reduction with $\mathrm{poly}(1/\eps)$ overhead.

\begin{theorem}[Main, Informal]\label{thm:main-informal}
Agnostic learning and multiaccuracy are statistically and computationally equivalent in the sense that for all feature distributions $\D_\X$ and model classes $\C$:
\begin{itemize}
    \item There exists a $(\C, \eps)$-agnostic learning algorithm for $\D_\X$ that only accesses the data distribution via $O(1/\eps^5)$ calls to a $(\C, \Omega(\eps^2))$-multiaccuracy oracle for $\D_\X$.
    \item (Conversely, there exists a $(\C, \eps)$-multiaccuracy algorithm for $\D_\X$ that accesses the data distribution via $O(1/\eps^2)$ calls to a $(\C, \Omega(\eps))$-agnostic learning oracle for $\D_\X$ \cite{TrevisanTV09, hkrr2018}.)
\end{itemize}
\end{theorem}

The converse direction of this equivalence was previously known. Indeed, the powerful \emph{regularity lemma} of \cite{TrevisanTV09} states that efficient $(\C, \eps)$-multiaccurate models always exist in the form of clipped linear combinations of $O(1/\eps^2)$ functions from $\C$,\footnote{The original versions of the lemma pertained to the specific class $\C$ of \emph{cuts} in graphs \cite{szemeredi1975regular,frieze1996regularity}.} and it was later shown that these component functions can be efficiently found using adaptive calls to a (weak) agnostic learning oracle \cite{hkrr2018}.\footnote{Technically, the $O(1/\eps^2)$ summands found by the oracle need not belong to the class $\C$, unless we happen to be working with a \emph{proper} (weak) agnostic learning oracle.} Agnostic learning turns out to be a simple yet powerful learning primitive.
Other examples of goals that are known to be achievable using an agnostic learning oracle include \emph{multicalibration}, a stronger version of \eqref{eq:multiaccuracy-intro} that conditions on the level sets of $p$, and \emph{omniprediction}, a stronger version of \eqref{eq:agnostic-learning-intro} that requires small error not just for the $\ell_1$ loss but also simultaneously for a multitude of other loss functions in a set $\L$ \cite{hkrr2018,omni}. Provided $\ell_1 \in \L$, omniprediction implies agnostic learning. 

The forward direction was open prior to our work. Positive answers were known under stronger assumptions on the predictor, typically that it satisfies calibration guarantees \cite{dawid1985calibration} of varying strengths. It was known that if a model is both $\C$-multiaccurate and calibrated, then it must be an agnostic learner for the class $\C$ \cite{CGKR26}. The stronger guarantee of multicalibration was known to imply omniprediction for a rich family of convex losses \cite{omni}, and hence agnostic learning. These assumptions can be viewed as requiring stronger indistinguishability guarantees between the real and simulated data distributions, via the outcome indistinguishability framework \cite{OI}.

The above reductions to agnostic learning were one-shot reductions, where we take the predictor $p$ satisfying calibrated multiaccuracy, and output $\sign(p(x))$ as our learner. It was shown by \cite{CGKR26} that such reductions will not suffice for multiaccuracy: there are multiaccurate predictors where even {\em conditioned on the output} of the predictor, the true label is uniformly random (a condition they called \emph{anti-calibration}). This implies that no post-processing of the predictor can beat random guessing.  But it leaves open the possibility that a more sophisticated reduction to multiaccuracy (for example, with repeated oracle calls) could still yield agnostic learning. Note that the reduction in the other direction crucially uses multiple rounds with adaptive calls to the agnostic learner to get a multiaccurate predictor.

In this work, we show that access to a \emph{multiaccuracy oracle} for a class $\C$ is indeed sufficient for agnostic learning of $\C$. We prove this by designing a statistically and computationally efficient oracle reduction from agnostic learning to multiaccuracy. By ``multiaccuracy oracle'', we mean an algorithm that receives as input $n$ samples from an arbitrary labeled data distribution with fixed marginal $\D_\X$ over features and outputs a predictor that is $(\C, \eps)$-multiaccurate with respect to this distribution. We find this main result surprising given the aforementioned negative results regarding the post-processing of a single multiaccurate predictor. 
Moreover, our reduction holds even in the statistical query (SQ) framework of \cite{kearns98sq}. This follows by standard arguments for implementing boosting-style procedures using statistical queries \cite{feldman09complete}. Importantly, this implies that SQ lower bounds for agnostic learning also extend to learning multiaccurate predictors. See the related work section, as well as \Cref{appendix:sq-preservation}, for more details.

\subsection{Why This Direction Is Harder}

The standard reduction from multiaccuracy to weak agnostic learning can be seen as a boosting algorithm that proceeds in rounds: at round $t$, assuming we have not achieved multiaccuracy, we use the weak learner to find $c_t$ that is correlated with the residual error $\y - p_t(\x)$ of our predictor. We think of $(\y - p_t(\x))/2$ as the expectation of a Bernoulli random variable $\z$, which $c_t$ is guaranteed to predict better than random guessing. We update our predictor using $c_t$ so as to reduce the squared error. A standard potential argument shows that this cannot proceed for too many rounds. 

Why does boosting or some similar procedure not work in the reverse direction? It is because we work with a different primitive: At stage $t$, we are given a $\C$-multiaccurate predictor $p_t$, which knows all the structure in the target distribution (that it was trained on) that is observable to the class $\C$. Yet, the key challenge is that $p_t$ need not be any better at predicting $\y$ than random guessing. The following example illustrates this point. 
\paragraph{A Guiding  Example.}
\label{sec:symmetry-example}

Consider a data distribution with uniform $n$-bit features $\rn{x} \sim \{\pm 1\}^n$ and binary labels $\rn{y} = \rn{x}_n$. Let $\C$ be the class of \emph{symmetric} functions $c : \{\pm 1\}^n \to [\pm 1]$, which are invariant under permutations of the input bits. In this setting, there are many perfectly multiaccurate predictors. For example, consider the predictors $p_i(x) = x_i$ for each index $i \neq n$.
By symmetry of the $i$\textsuperscript{th} and $n$\textsuperscript{th} bits, $p_i$ satisfies $(\C, 0)$-multiaccuracy. In fact, it does so even after conditioning on the predictions made by $p_1, \ldots, p_{i-1}$:
\[
    \E\bigl[c(\rn{x})(\rn{x}_n - \rn{x}_i) \,\big|\, \rn{x}_1, \ldots, \rn{x}_{i-1}\bigr] = 0.
\]
Indeed, from the point of view of symmetric tests, $\x_i$ is a fine model for $\y = \x_n$. 
However, since the $n$ coordinates of $\rn{x}$ are independent, the predictors $\x_1, \ldots, \x_{n-1}$ are essentially random guesses for the label! 

Assume that our goal is to achieve squared loss comparable to the best symmetric function. Each $p_i$
gives a squared loss of $2$ since
\[ \E[(\y - \x_i)^2] = 2 - 2\E[\x_i\x_n] = 2. \]
In contrast, the best symmetric function $s_n(\x) = \sum_i x_i/n$ achieves error $1  - 1/n$:
\begin{align}
    \label{eq:sn-err}
    \E\bigl[(\rn{y} - s_n(\rn{x}))^2\bigr] = \E\biggl[\Bigl(\rn{x}_n - \frac{\rn{x}_1 + \cdots + \rn{x}_n}{n}\Bigr)^2\biggr] = 1 - \frac{1}{n}.
\end{align}

Now assume that $n$ is very large compared to the number of queries $t$ we can make to the multiaccuracy oracle. Given $\x_1, \ldots, \x_t$ for $t \ll n$ in response to our queries, can we combine these random guesses to improve on the squared loss of each individual guess? The answer is yes: we output the average $s_t(\x) = \sum_{i \leq t}\x_i/t$, whose squared error is seen to be bounded by $1 + 1/t$. Hence for $t = O(1/\eps)$, we are within $1/n + 1/t = O(\eps)$ of the optimal squared error, since $t < n$. 

What can we learn from this? Note that $s_t(\x)$ is also independent of $\x_n$, just like each $\x_i$ for $i \leq t$. It is also $\C$-multiaccurate, being the average of multiaccurate predictors. However, unlike the $\x_i$s and like $s_n$, it has {\em low calibration error}. By the result of \cite{gopalan2024swap, CGKR26}, multiaccuracy together with calibration implies agnostic learning. So one lesson is to try and make calibrated predictions, while not losing multiaccuracy.

Another key idea comes from considering the agnostic learning setting, where the goal is to achieve correlation comparable to the best hypothesis in $\C$.
As \cite{CGKR26} showed, there are distributions $\D$ and classes $\C$  where the optimal hypothesis from the class $c$ satisfies $\E[c(\x) \y] \geq 1/2$, whereas there exist multiaccurate predictors where $\E[p(\x)\y] = 0$; in fact the stronger anti-calibration condition $\E[\y|p(\x)] = 0$ holds. This anti-calibration condition in their example also shows that na\"{i}vely attempting to calibrate a multiaccurate predictor doesn't work as we simply get the constant zero function after calibrating. We can reconstruct a similar example by fixing an odd dimension $n \geq 3$, setting $\y = \x_n$, and letting $\alpha := \E[\mathrm{Maj}_n(\x)\y] = \Theta(1/\sqrt{n})$, where $\mathrm{Maj}_n$ is the Majority function on $n$ variables.
Then, $\mathrm{Maj}_n$ achieves correlation $\alpha$, which is optimal for $\C$ (all symmetric functions).
But every convex combination of $\x_1,\ldots, \x_{n-1}$ is perfectly $\C$-multiaccurate, yet has $0$ correlation with $\x_n$. Our aim is to learn a hypothesis with correlation close to $\alpha$. Clearly, we cannot achieve this by averaging predictors that have $0$ correlation with $\y$. So we need the multiaccuracy oracle to not just repeat combinations of $\x_1, \ldots, \x_{n-1}$, but to eventually reveal $\x_n$. In general, we want to extract from the oracle a diverse set of predictors, including ones that have non-trivial correlation. 
These observations inspire our main reduction, which we describe in the next section.

\subsection{Algorithm Overview}

\paragraph{High-Level Outline.} We query the multiaccuracy oracle to get predictors $p_1, \ldots, p_t$ that are all $\C$-multiaccurate for the target distribution $\D$. At each stage, we ensure that the predictor $p_t$ is {\em weakly independent} of the predictors seen so far. We then output a randomization/weighted average of the various $p_t$ as our final predictor. If we think of each $p_t$ as having a structured component along $\C$ and an orthogonal noise component, then averaging preserves the structure but the noise cancels out. Over multiple rounds, we prove that randomizing over the predictors yields a predictor that is both multiaccurate and calibrated, and this implies agnostic learning. Indeed, this approach naturally yields the much stronger primitive of swap multicalibration, which is known to guarantee swap omniprediction, a strong generalization of agnostic learning \cite{gopalan2024swap}.  

The question that remains is finding the right notion of {\em weak independence}. We require that the predictor $p_t$ is multiaccurate conditioned on the level sets of $\sum_{i <t}w_ip_i/(\sum_{i<t} w_i)$, suitably discretized, for some suitably chosen weights $w_i \geq 0$. We show how to obtain such a predictor efficiently using calls to a standard multiaccuracy oracle, changing only the conditional distribution over the labels $\y$, while leaving the marginal distribution over $\rn{x}$ fixed. By stitching together the multiaccurate predictors for each level set, we ensure that the overall predictor is multiaccurate for the target distribution $\D$ that we wish to learn agnostically. 

The analysis of this algorithm is fairly subtle; it combines ideas from classic algorithms like Follow-the-leader in the online learning literature \cite{Hannan1957}, with the technique of loss outcome indistinguishability from more recent work on omniprediction \cite{lossOI}. Randomization over the predictors $p_i$, which is typical in online learning, is in fact essential for achieving strong notions like omniprediction or multicalibration that give guarantees for multiple losses or distinguishers, since different $p_i$s might do well for different losses. In contrast, for agnostic learning (single loss) or degree-$2$ calibration (single distinguisher), one can pick the single best predictor over the iterations. 

\medskip
We now present a more formal outline of the reduction 
from agnostic learning to multiaccuracy, which has several steps. It is simple to state what each step aims to accomplish;  however the analysis is far from obvious. For this reason, we first explain the steps of the algorithm and defer all analysis to the technical overview in \cref{sec:tech}.

\paragraph{Step 1: Reweighting.}

\begin{algorithm}[t]
\caption{Weighted Multiaccuracy}
\label{alg:reweight}

\begin{algorithmic}[1]
    \Require Multiaccuracy oracle $\textsf{MA}$ with sample size $n$, reweighting measure $m : \X \to [0, 1]$, labeled data points $(x_{ti}, y_{ti}) \in \X \times \{\pm 1\}$ for $t \in [T]$ and $i \in [n]$.

    \Statex
    
    \State $p_0 = 0$ \Comment{Learning}

    \For{$t = 1, \ldots, T$}
        \State Sample $\rn{z}_{ti} \in \{\pm 1\}$ with
        $\E[\rn{z}_{ti} | x_{ti}, y_{ti}] = m(x_{ti})\cdot y_{ti} + (1-m(x_{ti}))\cdot p_{t-1}(x_{ti})$
        \State $p_t = \textsf{MA}((x_{ti}, \rn{z}_{ti})_{i \in [n]})$
    \EndFor

    \Statex

    \State \Return $p = \frac{1}{T}\sum_{t=1}^T p_t$
    \Comment{Averaging}

\end{algorithmic}
\end{algorithm}

The first step, described in \cref{alg:reweight}, converts a $(\C, \eps)$-multiaccuracy oracle for a fixed feature distribution $\D_\X$ into a multiaccuracy oracle for any reweighting of $\D_\X$. The algorithm takes as input a function $m : \X \to [0, 1]$, which we view as a \emph{measure} that is allowed to decrease (but not increase) the weight of some feature vectors $x \in \X$. If we seek multiaccuracy conditional on the level sets of an arbitrary function $f : \X \to \V$ of the features, then running \cref{alg:reweight} on each of the $\{0,1\}$-valued level-set indicator functions $m = \bm{1}_{f^{-1}(v)}$ achieves the following guarantee, which we call \emph{$(\C, \eps)$-conditional multiaccuracy}.

\begin{theorem}[Conditional Multiaccuracy]
\label{thm:cond}
    For all $f : \X \to \V$, running \cref{alg:reweight} for $T = O(1/\eps)$ rounds with a $(\C, \eps)$-multiaccuracy oracle on each level set of $f$ yields a predictor $p : \X \to [\pm 1]$ satisfying\scmnote{Shouldn't we also add running Alg. 1 ``for $T=O(1/\epsilon)$ rounds''?}
    \[
        \E\biggl[\max_{c \in \C}\, \Bigl\lvert \E\bigl[c(\rn{x})(\rn{y} - p(\rn{x})) \mid f(\rn{x})\bigr]\Bigr\rvert\biggr] \le O \left(\abs{\V}\eps\right).
    \]
    If the oracle has failure probability $\delta$, then the resulting Conditional Multiaccuracy Algorithm has failure probability at most $O\left(\abs{\V}\delta/\eps\right)$.\vkmnote{I changed the eqn. above to add $O(.)$; also I think the failure probability is $T|\V|\delta$ where $T = O(1/\eps)$? I've made the change here.}
\end{theorem}

\cref{alg:reweight} operates by making a sequence of queries to the (unweighted) multiaccuracy oracle, using the same feature distribution but different label distributions at each time step. Specifically, at time $t \le T$, our algorithm uses a label distribution that preserves the true label of $x \in \X$ with probability $m(x)$, and otherwise replaces it with the prediction $p_{t-1}(x)$ of the oracle's previous output; since binary labels are required, we randomly round the prediction to $\{\pm 1\}$ accordingly.

\paragraph{Step 2: Agnostic Learning.}

\begin{algorithm}[t]
\caption{Agnostic Learning}
\label{alg:main-al}

\begin{algorithmic}[1]
    \Require Conditional multiaccuracy oracle $\textsf{CMA}$, grid size $\gamma^{-1} \in \N$, function $f : [\pm 1] \to [\pm B]$

    \Statex
    
    \For{$t = 1,2,\ldots,T$}

        \State $q_t = \sum_{s<t} (s \cdot r_s)/\binom{t}{2}$ \Comment{Response aggregation ($q_1=0$)}
        
        \State $r_t = \textsf{CMA}(R_\gamma \circ f \circ q_{t})$ \Comment{$R_\gamma$ rounds to multiples of $\gamma$}

    \EndFor
     \Statex

    \State $p = \sum_{t=1}^T (t \cdot (f \circ q_t)) / \binom{T+1}{2}$ \Comment{Averaging}

    \Return $p$

\end{algorithmic}

\end{algorithm}

The next part of our reduction, described in \cref{alg:main-al}, also makes iterative calls to an oracle, but this time to a \emph{conditional} multiaccuracy oracle, such as the one in \cref{thm:cond}. The following theorem states the agnostic learning guarantee that it achieves.

\begin{theorem}[Agnostic Learning]
\label{thm:main-al}
    \cref{alg:main-al} with a $(\C, \gamma\eps)$-conditional multiaccuracy oracle, soft-thresholding $f=\sign_\eps$, grid size $\gamma = \eps$, and time horizon $T = O(1/\eps^2)$ outputs $p$ satisfying
    \[
        \E\bigl[\lvert \rn{y} - p(\rn{x})\rvert\bigr] \le \min_{c \in \C}\,\E\bigl[\lvert \rn{y} - c(\rn{x})\rvert\bigr] + O(\eps).
    \]
    If the oracle has failure probability $\delta$, then \cref{alg:main-al} has failure probability at most $T \delta$.
\end{theorem}

At each time step $t \le T$, \cref{alg:main-al} first computes a weighted average $q_{t} : \X \to [\pm 1]$ of the oracle responses $r_1, \ldots, r_{t-1}$ received so far, placing slightly more weight on more recent responses. It then post-processes $q_t$ with the carefully chosen soft-thresholding function
\[
    f(v) = \sign_\eps(v) =  \begin{cases}
            -1 &\text{if }v \le -\eps,\\
            +1 &\text{if }v \ge +\eps,\\
            v/\eps &\text{otherwise.}
        \end{cases}
\]
Next, the algorithm queries the conditional multiaccuracy oracle on $f \circ q_t$, rounded to multiples of $\gamma$, to receive a new response $r_t$. Finally, the algorithm outputs a nonuniform weighted average of $f \circ q_t$ over time steps $t \in [T]$.

Note that Theorem \ref{thm:main-al} implies that if the optimal correlation with $\C$ is a constant $\alpha$, then the algorithm gets correlation $\alpha/2$ in $O(1/\alpha^2)$ rounds, hence in constantly many rounds. In our guiding example with $n = \Theta(1/\alpha^2)$ variables, it implies that within $O(n)$ rounds, the conditional multiaccuracy oracle must return a predictor that depends on $\x_n$. 

Note that the averaging over stopping times $t$ is a strength of the algorithm, not a weakness. Essentially, our guarantee states that the post-processed iterates $q_t$ perform well on average. In particular, if one were to output the one with the lowest $\ell_1$ loss, which can be identified with a handful of additional labeled samples, this would also perform well. We also remark that the nonuniform weighting in both the response aggregation and final averaging is not strictly necessary; as our proof will show, if we were to use a uniform weighting for both the average over $r_t$ in the inner loop and the average over $f \circ q_t$ at the end, it would only cost an additional factor of $\log(1/\eps)$ in the time horizon $T$.

\paragraph{Step 3: Omniprediction and Multicalibration.}

\begin{algorithm}[t]
\caption{Multicalibration and Omniprediction}
\label{alg:main}

\begin{algorithmic}[1]
    \Require Conditional multiaccuracy oracle $\textsf{CMA}$, grid size $\gamma^{-1} \in \N$, learning weights $a_1, \ldots, a_T \ge 0$, mixing weights $b_1, \ldots, b_T \ge 0$, noise function $\rn{N} : [\pm 1] \to [\pm 1]$.

    \Statex
    \For{$t = 1,2,\ldots,T$}
        \State $q_t = (a_1r_1 + \cdots + a_{t-1}r_{t-1})/(a_1 + \cdots + a_{t-1})$ \Comment{Response aggregation ($q_1=0$)}
        \State $r_t = \textsf{CMA}(R_\gamma \circ q_{t})$ \Comment{$R_\gamma $ rounds to multiples of $\gamma$}

    \EndFor
    \Statex

    \Function{$\rn{p}$}{$x$} \Comment{Randomization}
        \State $\rn{t} \sim b$
        \State \Return $\rn{N}(q_{\rn{t}}(x))$
    \EndFunction

    \Statex
    \State \Return $\rn{p}$

\end{algorithmic}
\end{algorithm}
\pgmnote{Might be worth working out what the standard conversion to omniprediction using calls to an agnostic learner gives for multicalibration. We should point out that the gains are pretty substantial.}
At this point, we have shown that multiaccuracy is sufficient to achieve agnostic learning, which was previously known to be sufficient to achieve strong learning-theoretic goals like \emph{omniprediction} and \emph{multicalibration}. Our next result shows that with minor modifications to \cref{alg:main-al}, we can actually achieve direct conversions from multiaccuracy into omniprediction and multicalibration that are much more efficient than going through agnostic learning as an intermediate step and then composing with known reductions.
As a concrete comparison, for the case of multicalibration, the algorithm of \cite{omni} uses $O(1/\epsilon^{6})$ calls to a weak agnostic learner. 
Each of these calls can in turn be implemented at the required accuracy using $O(1/\epsilon^{2})$ conditional multiaccuracy calls, by our Theorem~\ref{thm:main-al}.
This composition therefore uses a total of $O(1/\epsilon^{8})$ conditional multiaccuracy calls to achieve multicalibration.
In comparison, our Theorem~\ref{thm:main} below achieves multicalibration (and the stronger guarantee of swap multicalibration) at the same accuracy using only $O(1/\epsilon^{3})$ conditional multiaccuracy calls.

The notion of omniprediction we consider is with respect to a class of loss functions that we call \emph{naturally parametrized} losses. To define this class, we need some notation. Consider a real-valued loss function $\ell(y, a)$, which takes as input a label $y \in \{\pm 1\}$ and an action $a$ in some arbitrary action set. Given a value $p \in [\pm 1]$, which we identify with the unique probability distribution over $\{\pm 1\}$ with mean $p$, we define
\begin{align*}
    \ell(p, a) &= \E_{\rn{y} \sim p}[\ell(\rn{y}, a)], &\text{(expected loss)} \\
    k_\ell(p) &\in \argmin_{a}\, \ell(p, a). &\text{(best response)}
\end{align*}
Note that $k_\ell(p)$ does not necessarily exist for an arbitrary loss $\ell$ and value $p \in [\pm 1]$, and even when it does exist, it might not be unique. We also define the discrete partial derivative
\[
    \partial \ell(\eta) = \frac{\ell(+1, \eta) - \ell(-1, \eta)}{2}.
\]
We say that $\ell$ is \emph{naturally parametrized} if its action set is $\R$ and $\partial \ell(\eta) = -\eta$. This class includes, in particular, all \emph{generalized linear model (GLM)} loss functions, a class of loss functions that are convex in this natural parameter, which have also arisen in prior work on omniprediction \cite{lossOI}, but the losses we consider need not be convex. We let $\L_B$ be the class of naturally parametrized loss functions with bounded best responses $k_\ell(p) \in [\pm B]$ and we let $\L_{B,\beta} \subseteq \L_B$ further require that $k_\ell$ is $\beta$-Lipschitz. For examples of such loss functions, see \cref{sec:preliminaries}.

\cref{alg:main} differs from \cref{alg:main-al} in a few ways. Now, at step $t$, we split the domain by the level sets of $q_{t}$, which is loss-independent, as opposed to splitting by the level sets of $f \circ q_t$ for some loss-dependent post-processing $f$. It also outputs a \emph{randomized} predictor derived from the entire sequence of iterates $q_1, \ldots, q_T$, rather than any particular chosen $q_t$ from the sequence. We also sometimes add additional noise.\footnote{Note that the addition of noise in \cref{alg:main} may be effectively omitted by plugging in the identity function, which in particular is deterministic, for the noise function parameter $\rn{N}$.} \cref{thm:main} below summarizes the various learning guarantees attained by \cref{alg:main}. We aim for omniprediction with respect to all suitably bounded naturally parametrized loss functions, with better rates when $k_\ell$ is Lipschitz. We also target the challenging ``swap'' versions of multicalibration, along with a novel ``smooth swap'' version. We defer the formal definitions of omniprediction and multicalibration to the next section.

\begin{theorem}
\label{thm:main}
    \cref{alg:main} with a $(\C, \gamma\eps)$-conditional multiaccuracy oracle outputs a randomized predictor with any one of the learning guarantees listed below, for some appropriate choices of the algorithm's parameters $a_t$, $b_t$, and $\rn{N}$ (the choices depend on the particular guarantee):
    {
        \setlength{\tabcolsep}{2pt}
            \begin{longtable}{llll} 
                $(\C, O(\eps))$-smooth & omniprediction, & using $T = O(1/\eps)$ & oracle calls with $\gamma = \Omega(\eps)$; \\
                $(\C, O(\eps))$-  & omniprediction, & using $T = O(1/\eps^2)$ & oracle calls with $\gamma = \Omega(\eps^2)$;
                \vspace{1em} \\
                $(\C, O(\eps))$-smooth swap & multicalibration, & using $T = O(1/\eps^2)$ & oracle calls with $\gamma = \Omega(\eps)$; \\
                $(\C, O(\eps))$-swap & multicalibration, & using $T = O(1/\eps^3)$ & oracle calls with $\gamma = \Omega(\eps^2)$. \\
            \end{longtable} 
    }
    In each case, the predictor is also $(\C, O(\eps))$-multiaccurate. Omniprediction is with respect to $\L_1$ or $\L_{1,1}$. If the oracle has failure probability $\delta$, then \cref{alg:main} has failure probability at most $T\delta$.
\end{theorem}

We prove \cref{thm:main} in \cref{sec:upper}, where we also prove certain quantitative improvements in the case of a single loss function $\ell$ (not necessarily $\ell_1$). We show in \cref{sec:lower} that many of the rates in \cref{thm:main} are tight for our algorithm by constructing several specific hard inputs. We remark that omniprediction also has a challenging ``swap'' variant in the literature, which \cref{thm:main} does not explicitly attempt to achieve. However, it was shown in \cite{gopalan2024swap} that swap omniprediction is equivalent up to a small change of parameters to swap multicalibration, which our algorithm indeed achieves.\footnote{In fact, if we only allow swapping of the function $c$ (but not the loss $\ell$), then we can achieve this with the same error rates as for the non-swap versions laid out in \cref{thm:main}. Allowing swapping of both $c$ and $\ell$ incurs a larger error rate, owing to our larger error rate for swap multicalibration and the conversion to swap omniprediction.}

\subsubsection{Additional Definitions}

We now formally define multicalibration \cite{hkrr2018,gopalan2024swap} and omniprediction \cite{omni}. As discussed, multiaccuracy is a notion of indistinguishability, and multicalibration strengthens it by applying it to each level set of the predictor. When the class $\C$ contains only constant functions, multicalibration collapses to the much older notion of \emph{calibration} \cite{Dawid}. In some cases, we allow a different distinguisher from $\C$ to be applied to each level set, which we call \emph{swapping}. In our proofs, the notion of smoothness will play a central role, which is why we present a new notion of \emph{smooth swap multicalibration} alongside the existing notions. For a unified definition, we use the framework of \emph{sample-access outcome indistinguishability (OI)} \cite{OI}.

\begin{definition}[Multiaccuracy and Multicalibration \cite{hkrr2018,gopalan2024swap}]
\label{def:multiaccuracy}
    Given a function class $\C \subseteq \{\X \to [\pm 1]\}$, let $\C^{\mathrm{swap}}$ be the class of all functions $c : [\pm 1] \times \X \to [\pm 1]$ such that $c_v(\cdot) = c(v,\cdot) \in \C$ for every value $v \in [\pm 1]$. A predictor $p : \X \to [\pm 1]$ is \emph{$(\C, \eps)$-swap multicalibrated} with respect to a distribution $\D$ over labeled data points $(\rn{x}, \rn{y}) \in \X \times \{\pm 1\}$ if for all $c \in \C^{\mathrm{swap}}$,
    \begin{equation}
    \label{eq:multi-x}
        \E\bigl[c_{p(\rn{x})}(\rn{x})(\rn{y}-p(\rn{x}))\bigr] \le \eps.
    \end{equation}
    There are several special cases, depending on whether or not $c$ is \emph{$L$-Lipschitz}, meaning $c(v, x)$ is $L$-Lipschitz in $v$ for every $x \in \X$, and whether or not $c$ \emph{factors}, meaning $c(v,x) = w(v)c'(x)$ for some $w : [\pm 1] \to [\pm 1]$ and some $c' \in \C$. If \eqref{eq:multi-x} holds...
    {
        \setlength{\tabcolsep}{2pt}
        \begin{table}[H]
            \centering
            \begin{tabular}{ll}
                ...when $c$ is $1$-Lipschitz,
                 & then $p$ is \emph{$(\C, \eps)$-smooth swap multicalibrated}; \\
                ...when $c$ factors,
                 & then $p$ is \emph{$(\C, \eps)$-multicalibrated}; \\
                ...when $c$ is $1$-Lipschitz and factors,
                 & then $p$ is \emph{$(\C, \eps)$-smooth multicalibrated}; \\
                ...when $c$ ignores its first input,
                 & then $p$ is \emph{$(\C, \eps)$-multiaccurate}. \\
            \end{tabular}
        \end{table}
    }
    
    We only consider base classes $\C$ that contain the constant $1$ function and are closed under negation and convex combinations. We omit the prefix ``multi'' when $\C$ only contains the constant functions (e.g., \emph{$\eps$-calibration} instead of $(\C, \eps)$-multicalibration or $(\C, \eps)$-swap multicalibration, and \emph{$\eps$-smooth calibration} instead of $(\C, \eps)$-smooth multicalibration or $(\C, \eps)$-smooth swap multicalibration). When $\rn{p}$ is a randomized predictor, the expectation in \eqref{eq:multi-x} is also taken over its internal randomness.
\end{definition}

We remark that the definition of smooth swap multicalibration may be less intuitive than the others since it may be harder to imagine natural examples of functions $c \in \C^{\mathrm{swap}}$ that are Lipschitz but do not factor. For an example of where such functions naturally arise, one might consider the expected advantage of distinguishers $c({\rn{v}}, \rn{x})$ that observe only an unbiased noisy estimate $\rn{v}$ of $p(\rn{x})$, rather than the prediction $p(\rn{x})$ itself.

We now turn our attention to loss minimization and omniprediction \cite{omni}, with an emphasis on naturally parametrized loss functions. While most parts of the following definition are standard in the literature, the smooth variant is new.

\begin{definition}[Omnipredictor \cite{omni}]
    Given a function class $\C$ and a class of loss functions $\L$, a predictor $p$ is an \emph{$(\L, \C, \eps)$-omnipredictor} if for all $c \in \C$ and $\ell \in \L$,
    \begin{equation}
    \label{eq:def-op}
        \E\bigl[\ell(\rn{y}, k_{\ell}(p(\rn{x})))\bigr] \le \E\bigl[\ell(\rn{y}, c(\rn{x}))\bigr] + \eps.
    \end{equation}
    If inequality \eqref{eq:def-op} only holds when the function $\partial \ell(k_\ell(v))$ is $1$-Lipschitz, then we instead say that $p$ is an \emph{$(\L, \C, \eps)$-smooth omnipredictor}.
    
    By default, in this paper, $\L$ refers to the class of naturally parametrized loss functions, meaning that $\partial \ell(\eta) = -\eta$, with $[\pm 1]$-valued best responses $k_\ell(p)$. When this is clear from context, we simply say that $p$ is a \emph{$(\C, \eps)$-omnipredictor} or \emph{$(\C, \eps)$-smooth omnipredictor}.
\end{definition}

\subsection{Analysis Overview}
\label{sec:tech}

In this section, we present the ideas behind the analysis of our main oracle reductions from agnostic learning (and beyond) to multiaccuracy. There are multiple ways to frame our arguments. The most natural approach is perhaps through the lens of online learning, even though we study the batch setting. To begin, let us consider the following simple game played repeatedly between two players, named the \emph{Predictor} and \emph{Nature}.
\begin{definition}[Deterministic Online Prediction Game] At each time step $t = 1, \ldots, T$,
\begin{enumerate}[1.]
    \item The Predictor selects and announces a prediction $q_t \in [\pm 1]$. We require $q_t = s(r_1, \ldots, r_{t-1})$ to be a deterministic function $s$ of Nature's previous responses $r_1, \ldots, r_{t-1}$. We call the function $s$ the Predictor's \emph{strategy}.
    
    \item After seeing $q_t$, Nature adaptively selects and announces its response $r_t \in [\pm 1]$.
\end{enumerate}
\end{definition}

Ultimately, the Predictor has one of two possible goals. The first possible goal of the Predictor is to minimize its \emph{external regret} for a class of loss functions $\L$, which measures its excess loss incurred with respect to the best fixed action $a \in [\pm 1]$ in hindsight:
\[
    R_\L(s) = \max_{\ell \in \L}\, \biggl( \sum_{t=1}^T \ell(r_t, k_\ell(q_t)) - \min_{a \in [\pm 1]} \sum_{t=1}^T \ell(r_t, a)\biggr).
\]

As always, we focus on loss functions $\ell(\cdot, \cdot)$ that are affine in their first input. We emphasize that both the Predictor's action and Nature's response are real values in the interval $[\pm 1]$, rather than discrete values in $\{\pm 1\}$.

The second possible goal for the Predictor is to minimize its \emph{calibration error} with respect to a class $\W$ of weight functions $w : [\pm 1] \to [\pm 1]$. Intuitively, online calibration asks that the error $r_t - q_t$ should be uncorrelated with $w(q_t)$. In the extreme, $\W$ contains all functions, in which case $\W$-calibration requires that $\E[r_{\rn{t}}|q_{\rn{t}}] = q_{\rn{t}}$ for a random time step $\rn{t} \sim \Unif\{1, \ldots, T\}$. For a given weight class $\W$, we define the calibration error as
\[
    \CE_\W(s) = \max_{w \in \W}\, \sum_{t=1}^T w(q_t)(r_t - q_t).
\]

\subsubsection{Multiaccuracy Oracle Reductions by Online Regret/Calibration}

The fundamental observation of our work is that any efficiently computable online strategy $s$ for the Predictor with low regret $R_\L(s)$ can be converted into a computationally efficient oracle reduction from $(\L, \C)$-omniprediction to $\C$-multiaccuracy. Similarly, if the Predictor's strategy achieves low online calibration error $\CE_\W(s)$, then we can convert it into an efficient oracle reduction from $\C$-swap multicalibration (for weight functions in $\W$) to $\C$-multiaccuracy.

Suppose we are given access to an oracle that takes as input a \emph{query} predictor $q : \X \to [\pm 1]$ and produces a \emph{response} $r : \X \to [\pm 1]$ that is multiaccurate on each level set of $q$. Let us call this conditional multiaccuracy oracle repeatedly across time steps $t = 1, \ldots, T$, where at each round $t$ we submit the following query, defined pointwise for each $x \in \X$ as
\[
    q_{t}(x) = s(r_1(x), \ldots, r_{t-1}(x)).
\]
If the Predictor has regret $R_\L(s) \le \delta T$, then applying this bound pointwise for each $x \in \X$ yields
\begin{equation}
\label{eq:tech-overview-bound}
    \sum_{t=1}^T \ell(r_t(x), k_\ell(q_{t}(x))) \le \min_{c : \X \to [\pm 1]} \sum_{t=1}^T \ell(r_t(x), c(x)) + \delta T.
\end{equation}
Note that since the bound has been applied pointwise for each $x \in \X$, we may choose a different fixed action $c(x)$ for each $x \in \X$ on the right-hand side.
Next, since $\ell$ is naturally parametrized, we have the special identity
\[
    \ell(r, \eta) - \ell(y, \eta) = \eta(y-r).
\]
This identity tells us that when the correlation of $\eta$ with $y-r$ is small, the action $\eta$ incurs similar loss on a real label $y \in \{\pm 1\}$ as on a simulated label generated from $r \in [\pm 1]$. By $(\C, \eps)$-conditional multiaccuracy,\footnote{In this technical overview, we write $\eps$ for the overall conditional multiaccuracy error across level sets, suppressing the discretization and per-level set accounting. What we really mean is $(\C, \gamma \eps)$-conditional multiaccuracy, applied to splitting functions with $1/\gamma$ level sets.} we know that simulated labels generated from the model $r_t$ are indistinguishable from real labels by both the entire class $\C$ and by any post-processing of $q_{t}$ \emph{in expectation} over $(\rn{x}, \rn{y})$. Accordingly, we may replace $r_t(\rn{x})$ with $\rn{y}$ in inequality \eqref{eq:tech-overview-bound} whenever $c \in \C$, which gives
\[
    \sum_{t=1}^T \E[\ell(\rn{y}, k_\ell(q_{t}(\rn{x})))] \le \min_{c \in \C} \sum_{t=1}^T \E[\ell(\rn{y}, c(\rn{x}))] + (2\eps+\delta)T.
\]
This is exactly $(\L, \C, 2\eps + \delta)$-omniprediction, not for any single predictor $q_{t}$ in the sequence, but rather 
for the \emph{randomized} predictor $\rn{p}$ that on input $x$ outputs $q_{\rn{t}}(x)$ for a freshly sampled time index $\rn{t} \sim [T]$. If the set $\L = \{\ell\}$ contains a single convex loss $\ell$, as is the case for agnostic learning, then this randomization is not even necessary, since we may simply output the average of the (post-processed) predictors $q_{t}$ in the sequence.

Similarly, if the Predictor has online calibration error $\CE_\W(s) \le \delta T$, then we get the following pointwise bound, where $c(v, x)$ is any function such that $c(\cdot, x) \in \W$ for all $x \in \X$:
\[
    \sum_{t=1}^T c(q_t(x), x)(r_t(x) - q_t(x)) \le \delta T.
\]
Once again, by applying $(\C, \eps)$-conditional multiaccuracy, we may replace $r_t(\rn{x})$ with $\rn{y}$ whenever $c(v, \cdot) \in \C$ for all values $v \in [\pm 1]$. This gives us either $(\C, \eps
+\delta)$-swap multicalibration or $(\C, \eps+\delta)$-smooth swap multicalibration, depending on whether the class $\W$ contains all possible bounded weight functions or just those bounded weight functions that are $1$-Lipschitz:
\[
    \sum_{t=1}^T \E[c(q_t(\rn{x}), \rn{x})(\rn{y} - q_t(\rn{x}))] \le (\delta + \eps)T.
\]

\subsubsection{Online Regret/Calibration by Following the Leader}

At this point, all that remains is to determine which loss classes $\L$ and which weight classes $\W$ admit efficient strategies for the Predictor with low regret and calibration error. For this, we use a classic and extremely simple algorithm in online learning, known as \emph{Follow the Leader (FTL)} (see \cite{Hannan1957,cesa2006prediction}). At each time step $t$, this algorithm makes a prediction that would have been the best fixed response in hindsight to the $t-1$ rounds completed so far.

In order to analyze this algorithm, it is helpful to introduce two quantities, which we call the \emph{entropy} and \emph{divergence}, that can be defined for any loss function $\ell$:
\begin{align*}
    h_\ell(p) &= \inf_{a} \, \ell(p, a), &\text{(entropy)} \\
    D_\ell(p, a) &= \ell(p, a) - h_\ell(p). &\text{(divergence)}
\end{align*}
The entropy is a concave function measuring the irreducible or inherent loss when the true distribution is given by $p$. The divergence is a nonnegative function measuring the excess loss of using a suboptimal action $a$.\footnote{For cross-entropy loss, $h_\ell$ and $D_\ell$ are the usual notions of \emph{Shannon entropy} and \emph{Kullback-Leibler (KL) divergence}. More generally, for \emph{proper losses} ($k_\ell(p) = p$), $D_\ell$ is a \emph{Bregman divergence} of $h_\ell$, and when $\ell$ is naturally parametrized and convex, $D_\ell$ is the \emph{Fenchel-Young divergence} of $h_\ell$.} The two parts of this decomposition of the loss into $\ell(p, a) = h_\ell(p) + D_\ell(p, a)$ are sometimes called \emph{aleatoric uncertainty} and \emph{epistemic uncertainty} \cite{ahdritz2025uncertainty}. For example, $\ell_1(y, \eta) = |y - \eta|$ has entropy
\(
    h_{\ell_1}(p) = 1 - |p|.
\)
For more examples, see \cref{sec:preliminaries}.

It turns out that FTL has $O(\log T)$ regret if the loss function's entropy $h_\ell$ is \emph{smooth}, which is roughly equivalent to the requirement that $\partial \ell \circ k_\ell$ is Lipschitz. In fact, using a weighted variant of FTL, one can improve this to $O(1)$ regret. For non-smooth entropy functions like $h_{\ell_1}$, we use a smoothing trick based on the addition of a small amount of random noise to transfer our results from the smooth-entropy case. For more details, see the proof of our main upper bound in \cref{sec:upper}.

We show that, conveniently, the FTL update rule also works for online smooth calibration, which we again transfer to full calibration with a noise trick. To our knowledge, this use case of Follow the Leader was not previously known. Concretely, we fix $0< \lambda \leq 1$ and have the Predictor play
\[
    q_t = (1-\lambda) q_{t-1} + \lambda r_{t-1}.
\]
By modifying the proof of the familiar regret bound for Follow the Leader, we show that this algorithm achieves small smooth calibration error:
\[
    \max_{\text{$1$-Lipschitz $w$}} \CE_w\bigl(\text{Follow the Leader}\bigr) = O{\Bigl(\frac{1}{\lambda} + \lambda T\Bigr)}.
\]
If $\lambda = 1/\sqrt{T}$, then the online error is $O(\sqrt{T})$. We emphasize that deterministic online algorithms for smooth calibration were previously known and that the $O(\sqrt{T})$ rate is known to be achievable \cite{foster2018smooth,qiao2024distance,arunachaleswaran2025elementary} and in fact optimal \cite{parikshitpersonal}. 
We choose to use Follow the Leader for its explicit and straightforward handling of real-valued responses $r_t \in [\pm 1]$ from Nature, where previous results generally considered only discrete $\{\pm1\}$-valued responses.

\subsubsection{Lower Bounds and Derandomization}

All of our upper bounds based on \cref{alg:main-al,alg:main} make $T = \mathrm{poly}(1/\eps)$ queries of granularity $\gamma$ to a $(\C, \gamma\eps)$-conditional multiaccuracy oracle, which is in turn implemented using $1/\gamma^2\eps$ calls to an ordinary $(\C, \gamma\eps)$-multiaccuracy oracle. Although these reductions are efficient, they raise the question of whether the number of oracles calls can be improved further.
In \cref{sec:lower}, we construct several instances of classes $\C$ and distributions over pairs $(\rn{x}, \rn{y})$ for which many calls to a conditional multiaccuracy oracle are required to achieve agnostic learning. Some of these instances are specific to our particular oracle algorithm, while others are not. For more detail, see the discussion in \cref{sec:lower} and \cref{thm:smce-lower,thm:ece-lower,thm:smce-mixture}.

In some cases, our oracle reduction outputs a randomized predictor, partly due to mixing across a sequence of predictors and partly due to explicit noise injection into the predictor's output. We show in \cref{sec:derand} that these predictors can be derandomized in certain cases, either by avoiding adding the noise to begin with (\cref{thm:degree-2-cal}) or by fixing the random coin tosses in advance (\cref{thm:main-derand}).

\subsection{Related Work}

As discussed, the notion of agnostic learning dates back to the work of \cite{haussler1992decision,kearns1994toward}. The related notion of weak agnostic learning, as well as techniques for agnostic boosting, play an important role in this literature \cite{ben2001agnostic,MansourM2002,KalaiMV08,kk09,feldman2009distribution}. In particular, these techniques arise in the proof of the complexity-theoretic regularity lemma, or equivalently, the construction of multiaccurate predictors, as discussed earlier \cite{TrevisanTV09,hkrr2018,kgz}. These works are also responsible for initiating the now quite substantial line of work on multicalibration. Especially relevant to us in this body of work are the notions of outcome indistinguishability \cite{OI} and omniprediction, which concerns the simultaneous optimization of multiple loss functions \cite{omni,lossOI,DworkLLT,GargJRR24,luo2025improved,DworkHIPT25,gibbs2025sample,OKK25}. Besides their relevance to omniprediction and other machine learning tasks, multiaccuracy and multicalibration have recently seen a surprisingly rich collection of applications to complexity theory and cryptography \cite{DworkLLT,hu2023generative,casacuberta2024complexity,marcussen2024characterizing, CGKR26, dwork2026efficient}.

Our work can be seen as the culmination of a line of research starting in \cite{hkrr2018, omni}, and continued in \cite{lossOI, gopalan2024swap, OKK25, DworkLLT, CGKR26, CGKRST26} that aims to understand the learning power implied by multigroup fairness notions of varying strength. For a fixed class $\C$, we know that multicalibration yields omniprediction \cite{omni, gopalan2024swap} for convex losses, while calibrated multiaccuracy yields agnostic learning \cite{lossOI, CGKR26}, both via one-shot reductions that post-process the predictions. One-shot multiaccuracy is a much weaker learning primitive, that does not yield (even) weak agnostic learning \cite{CGKR26}.  With oracle access to multiaccuracy and adaptive queries over multiple rounds, we show that one recovers the full power of omniprediction and multicalibration.  
\cite{globus2026augmented} recently proposed a ``feature-augmented'' multicalibration algorithm that avoids discretization during boosting, which uses a squared-loss minimization oracle over an augmented concept class. 
A call to such an oracle is stronger than a call to the multiaccuracy oracle we assume, given that any function which beats random guessing in
squared error already gives a weak agnostic learner \cite{CGKR26}. 

Turing formulates the imitation game as a game between a machine and a human, where winning the game means convincing the distinguisher that you are the human \cite{turing1950test}. Turing considers the possibility that a machine might win the imitation game using a policy that is different from imitating the behavior of a human. He considers this unlikely, and proceeds by explicitly assuming that the best strategy is indeed imitation. Recent studies cast doubt on this assumption, by showing that LLMs are sometimes more likely to be labeled human {\em than humans themselves}, which suggests that imitation is not the best strategy \cite{jones2026turing}. However, the name {\em imitation} suggests that perhaps a revised goal of the imitation game ought to be indistinguishability, for which there is no better strategy than imitation. The underlying question of what learning goals imitation achieves is what drives our work.

\paragraph{Computational Complexity of Multiaccuracy.}
Our main result (Theorem~\ref{thm:main-informal}) establishes that multiaccurate prediction and agnostic learning have essentially equivalent computational complexity in the fixed-distribution setting. One direction of this equivalence was established by \cite{hkrr2018}, who showed that agnostic learning implies multiaccurate prediction via an efficient reduction. Our converse reduction shows that computational lower bounds for agnostic learning also apply to multiaccurate prediction.

This equivalence is particularly informative in the statistical query (SQ) model \cite{kearns98sq}, where the complexity of agnostic learning is closely connected to polynomial approximation. The seminal work of \cite{KKMS08} showed that low-degree polynomial approximators under the underlying feature distribution yield efficient agnostic learning algorithms, extending the classical result of \cite{linial1993constant} for the realizable (noise-free) setting. Conversely, \cite{dachman2014approximate} established that polynomial approximation degree characterizes the SQ complexity of agnostic learning for a broad family of concept classes under the uniform distribution on the Boolean hypercube. This characterization was subsequently extended to Gaussian feature distributions by \cite{diakonikolas2021optimality}. Since our reduction can equally be implemented in the SQ framework, these characterizations now extend to multiaccurate prediction. For a short survey of quantitative bounds on the complexity of agnostic learning obtained by combining these characterizations with results from approximation theory, see, e.g., \cite[Appendix C]{stavropoulos2026computational}.

\subsection{AI Methodology}

Some steps of this paper's proofs were incrementally accrued through repeated rounds of interactive discussion between the authors and ChatGPT-5.6 Sol. The model suggested conditioning by level sets, which we do in Step 1, as well as preliminary versions of Step 2, and preliminary versions of the hard inputs constructed in \cref{sec:lower}. The model's provided version of Step 2 initially had some issues: it required exponentially many oracle calls, was specific to the $\ell_1$ loss, and had a somewhat opaque proof using a mysterious potential function. With further prompting, the model was able to bypass the first issue. The authors addressed the second issue with the introduction of Step 3, which used tools from convex analysis to generalize the proof to loss functions besides $\ell_1$ and also achieve omniprediction and swap multicalibration.  The authors addressed the third issue using a connection to online learning, which significantly shortened, simplified, and modularized the proof.  Similarly, the authors simplified and clarified the examples used in \cref{sec:lower}, the model's versions of which were more complicated than necessary (e.g. three-dimensional instead of two-dimensional).

\paragraph{Acknowledgements.}
KS would like to thank Arsen Vasilyan for useful discussions regarding the derandomization procedure presented in \Cref{sec:general-derand}.

\section{Preliminaries}
\label{sec:preliminaries}

Throughout this paper, we write variables in boldface to indicate that they are random variables, such as a random feature vector $\rn{x} \in \X$ or a randomized predictor $\rn{p} : \X \to [\pm 1]$. We use the abbreviations $\{\pm 1\} = \{-1, 1\}$ and $[\pm 1] = [-1, 1]$.

\begin{definition}[Convexity]
    A function $f : \R \to \R$ is \emph{convex} if
    \[
        f(\lambda x + (1-\lambda)y) \le \lambda f(x) + (1-\lambda) f(y)
    \]
    for all $x,y \in \R$ and $\lambda \in [0, 1]$. We say that $g : \R \to \R$ is a subderivative of $f$ if
    \[
        f(y) \ge f(x) + g(x)(y-x)
    \]
    for all $x, y \in \R$. Similarly, we say that $-f$ is \emph{concave} and that $-g$ is a \emph{superderivative} of $-f$.
\end{definition}

\begin{definition}[Bregman Divergence]
\label{def:bregman}
    Given a convex function $f$ and subderivative $f'$, let
    \[
        D_f(y, x) = f(y) - f(x) - f'(x)(y-x).
    \]
    Note that $f$ is $\beta$-smooth if and only if $D_f(y,x) \le (\beta/2)(y-x)^2$. If $h$ is a concave function, we sometimes write $D_h$ instead of $D_{-h}$ when the meaning is clear from context.\footnote{Classically, Bregman divergences are only defined for differentiable convex functions. Our extension beyond differentiable convex functions is quite natural, as long as one bears in mind that the notation $D_f$ obscures the fact that the definition depends not only on $f$ but also on the choice of a particular subderivative of $f$ at each point.}
\end{definition}

\begin{definition}[Conjugacy]
    The convex conjugate of a function $f : \R \to \R$ is
    \[
        \hat{f}(y) = \sup_x\, xy - f(x).
    \]
    It is known that the \emph{biconjugate} $\hat{\hat{f}}$ is the greatest convex lower-semicontinuous lower bound on $f$.
\end{definition}

\begin{definition}[Loss Function Miscellany]
    Given a loss function $\ell : \{\pm 1\} \times \A \to \R$, let
    \begin{align*}
        \ell(p, a) &= \E_{\rn{y} \sim p}[\ell(\rn{y}, a)], &\text{(expected loss)} \\
        k_\ell(p) &\in \argmin_{a}\, \ell(p, a), &\text{(best response)} \\
        h_\ell(p) &= \inf_{a} \, \ell(p, a), &\text{(entropy)} \\
        D_\ell(p, a) &= \ell(p, a) - h_\ell(p). &\text{(divergence)}
    \end{align*}
\end{definition}

\begin{example}[Entropy Functions]\;
    \begin{itemize}
    \item \textbf{(Absolute Loss)} The loss $\ell_1(y,\eta) = |y-\eta|$ has a piecewise linear entropy
    \[
        h_{\ell_1}(p) = 1 - |p|.
    \]
    \item \textbf{(Squared Loss)} The loss $\ell_2(y,\eta) = (y-\eta)^2/2$ has a quadratic entropy
    \[
        h_{\ell_2}(p) = \frac{1}{2}(1-p^2).
    \]
    Also, $D_{\ell_2}(p, \eta) = (p-\eta)^2/2$ is squared distance.
    \item \textbf{(Logistic Loss)} The loss $\ell(y, \eta) = \log(1+e^{-2y\eta})$ has (rescaled) Shannon entropy
    \[
        h_{\ell}(p) = \frac{1+p}{2}\cdot\log\Bigl(\frac{2}{1+p}\Bigr)+\frac{1-p}{2}\cdot\log\Bigl(\frac{2}{1-p}\Bigr).
    \]
\end{itemize}
\end{example}

\begin{definition}[Proper Loss \cite{mccarthy,savage1971elicitation}]
\label{def:proper}
    A loss function $\ell$ is \emph{proper} if $\A = [\pm 1]$ and
    \[
        \ell(p, q) \ge \ell(p, p)
    \]
    for all $p, q \in [\pm 1]$. Equivalently, there is some superderivative $h'_\ell$ of $h_\ell$ such that
    \[
        \ell(p, q) = h_\ell(q) + h_\ell'(q)(p-q).
    \]
    Equivalently, $D_\ell(p, q)$ is a Bregman divergence of $h_\ell$. Equivalently, $\ell(p, p) = h_\ell(p)$.
\end{definition}

\begin{definition}[Natural Parametrization]
    We let $\L_B$ be the class of loss functions $\ell$ with $\A = \R$ and $\partial \ell(\eta) = -\eta$ and $[\pm B]$-bounded best responses $k_\ell(p)$ for all $p \in [\pm 1]$. We let $\L_{B,\beta} \subseteq \L_B$ further require $k_\ell(p)$ to be $\beta$-Lipschitz in $p \in [\pm 1]$.
\end{definition}

\begin{example}[Natural Parametrization Examples]\;
\begin{itemize}
    \item \textbf{(Absolute Loss)} If we define $\ell_1(y, \eta) = |y - \eta|$ for $y \in \{\pm 1\}$, then for $\eta \in [\pm 1]$, we have
    \[
        \ell_1(p, \eta) = 1 - p\eta.
    \]
    This can be extended to all $\eta \in \R$ as follows:
    \[
        \ell(p, \eta) = \max(1, |\eta|) - p\eta.
    \]
    This extension is naturally parametrized and has $k_{\ell}(p) = \sign(p)$ as a $[\pm 1]$-valued best-response (note that $\sign(0) \in [\pm 1]$ may be defined arbitrarily).
    \item \textbf{(Squared Loss)} If we define $\ell_2(y, \eta) = (y - \eta)^2/2$ for $y \in \{\pm 1\}$, then
    \[
        \ell_2(p, \eta) = \frac{1+\eta^2}{2} - p\eta.
    \]
    This is naturally parametrized with $k_{\ell_2}(p) = p$ as a $[\pm 1]$-valued best response.
    \item \textbf{(Logistic Loss)} If we define $\ell(y, \eta) = \log(1+e^{-2y\eta})$ for $y \in \{\pm 1\}$, then
    \[
        \ell(p, \eta) = \log(e^\eta + e^{-\eta}) - p \eta.
    \]
    This is naturally parametrized with best-response
    \(
        k_\ell(p) = \tanh^{-1}(p),
    \)
    which can be made $[\pm B]$-valued by modifying $\log(e^\eta+e^{-\eta})$ to instead be linear with slope $-1$ for $\eta < -B$ and linear with slope $1$ for $\eta > B$.
\end{itemize}

\end{example}
\begin{definition}[Calibration Error \cite{dawid1985calibration,GopalanKSZ22}]
    Given a class of weight functions $\W \subseteq \{[\pm 1] \to [\pm 1]\}$, the weight-restricted calibration error of a predictor $p : \X \to [\pm 1]$ with respect to $\W$ (and a distribution over labeled data points $\rn{x},\rn{y}$) is
    \[
        \CE_\W(p) = \max_{w \in \W}\, \E[w(p(\rn{x}))(\rn{y}-p(\rn{x}))].
    \]
    When $\W$ contains only the identity function, we instead write $\CE_2$, called \emph{degree-2 calibration error}. When $\W$ contains all $1$-Lipschitz functions, we instead write $\smCE$, called \emph{smooth calibration error}. When $\W$ contains all functions, we instead write $\ECE$, called \emph{expected calibration error}.
\end{definition}

\begin{definition}[Pseudodimension  \cite{haussler1992decision}]
\label{def:pseudodim}
    The \emph{pseudodimension} of a class $\C \subseteq \{\X \to \R\}$ is the VC dimension of the following class over the domain $\X \times \R$:
    \[
        \bigl\{(x, y) \mapsto \bm{1}[c(x) > y] \,\big|\, c \in \C\bigr\}.
    \]
    
\end{definition}

\section{Upper Bounds}
\label{sec:upper}

In this section, we analyze our main oracle reduction from agnostic learning to multiaccuracy. As discussed, our reduction also achieves several nominally stronger guarantees, such as (swap) omniprediction and (swap) multicalibration, in one fell swoop. We follow the template outlined in the technical overview in \cref{sec:tech}.

\subsection{Reweighting and Conditioning}

As described in the introduction, the first step of the reduction, described in \cref{alg:reweight}, allows us to transform a multiaccuracy oracle for the feature distribution $\D_\X$ into a conditional multiaccuracy oracle with respect to an arbitrary function $f : \X \to \V$. \Cref{thm:cond} states the guarantee for this step of the algorithm. To prove it, we first prove the following, slightly more general theorem.

\begin{theorem}
\label{thm:reweight}
    \cref{alg:reweight} with a $(\C, \eps)$-multiaccuracy oracle and $T = O(1/\eps)$ outputs $p$ satisfying
    \[
        \max_{c \in \C}\, \Bigl\lvert \E\bigl[m(\rn{x})\cdot c(\rn{x})(\rn{y} - p(\rn{x}))\bigr]\Bigr\rvert \le O(\eps).
    \]
    If the oracle has failure probability $\delta$, then \cref{alg:reweight} has failure probability at most $T\delta$.
\end{theorem}

\begin{proof}
    At each step $t \le T$, the $(\C, \eps)$-multiaccuracy oracle returns $p_t$ satisfying
    \begin{align}
        \max_{c \in \C}\, \bigl\lvert \E[m(\rn{x})\cdot c(\rn{x})(\rn{y}-p_t(\rn{x}))] + \E[(1-m(\rn{x}))\cdot c(\rn{x})(p_{t-1}(\rn{x})-p_t(\rn{x}))] \bigr\rvert &\le \eps. \label{eqn:MAguarantee}
    \end{align}
    In particular, for any fixed $c \in \C$ and for every $t \leq T$, we have,
    \begin{align}
         \E[m(\rn{x})\cdot c(\rn{x})(\rn{y}-p_t(\rn{x}))] + \E[(1-m(\rn{x}))\cdot c(\rn{x})(p_{t-1}(\rn{x})-p_t(\rn{x}))]  \le \eps. \nonumber
         \intertext{Let $\bar{p}_T = (p_1 + \cdots + p_T)/T$. Taking the average of the above inequality over all time steps $t \le T$ causes the second term on the left-hand side to telescope. This gives}
         \E[m(\rn{x}) \cdot c(\rn{x})(\rn{y}-\bar{p}_T(\rn{x}))] + \frac{1}{T}\E[(1-m(\rn{x}))\cdot c(\rn{x})(p_{0}(\rn{x})-p_T(\rn{x}))] &\le \eps. \label{eqn:avgMAupper}
         \intertext{Likewise, using Eq.~\eqref{eqn:MAguarantee} in the other direction, we can conclude that for any $c \in C$,}
         \E[m(\rn{x}) \cdot c(\rn{x})(\rn{y}-\bar{p}_T(\rn{x}))] + \frac{1}{T}\E[(1-m(\rn{x}))\cdot c(\rn{x})(p_{0}(\rn{x})-p_T(\rn{x}))] &\ge -\eps. \label{eqn:avgMAlower}
    \end{align}
    As Eqns.~\eqref{eqn:avgMAupper} and~\eqref{eqn:avgMAlower} apply to each $c \in \C$, rearranging the terms and using the triangle inequality, we conclude that
    \[
        \max_{c \in \C}\, \bigl\lvert \E[m(\rn{x}) \cdot c(\rn{x})(\rn{y}-\bar{p}_T(\rn{x}))] \bigr\rvert \le \eps + \frac{1}{T}. \qedhere
    \]
\end{proof}

\begin{proof}[Proof of \Cref{thm:cond}]
    Apply \cref{thm:reweight} to each $m = \bm{1}_{f^{-1}(v)}$ for each $v \in f(\X)$.
\end{proof}

\subsection{Achieving Agnostic Learning}
\label{sec:main-al-glm-proof}

We now analyze our reduction from agnostic learning to (conditional) multiaccuracy, described in \cref{alg:main-al}. As explained in the technical overview, the key tools that we use are the bounds on the online regret and online calibration error of Follow the Leader. We also use some smoothing tricks along the way. We defer these helper lemmas on FTL and smoothing to \cref{sec:upper-lemmas}. We actually prove the following, stronger version of \cref{thm:main-al}, which handles not just $\ell_1$ loss but any particular naturally parametrized loss satisfying mild conditions:

\begin{theorem}
\label{thm:main-al-glm}
    If $\ell$ is naturally parametrized and convex in its second argument with an $O(1)$-bounded and $O(1)$-Lipschitz best response function $k_\ell$, then \cref{alg:main-al} with a $(\C, \gamma\eps)$-conditional multiaccuracy oracle, the function $f = k_\ell$, grid size $\gamma = \eps$, and $T = O(1/\eps)$ outputs $p$ satisfying
    \[
        \E[\ell(\rn{y}, p(\rn{x}))] \le \min_{c \in \C}\,\E[\ell(\rn{y}, c(\rn{x}))] + O(\eps).
    \]
    Even if $k_\ell$ is not $O(1)$-Lipschitz, a modified choice of $f$ and slightly longer sequence of $T = O(1/\eps^2)$ calls suffice to achieve the inequality above. In both cases, if the oracle has failure probability $\delta$, then \cref{alg:main-al} has failure probability at most $T\delta$.
\end{theorem}

In the statement of \cref{thm:main-al-glm}, we also assumed that the loss function is convex, but we emphasize that this is \emph{only} for simplicity. The two real assumptions are (1) natural parametrization and (2) bounded best responses. If the loss function satisfies these two assumptions but is nonconvex, the error rates in \cref{thm:main-al-glm} still hold verbatim, but for a modified version of the algorithm that replaces the final averaging over iterates $t \in [T]$ with a random mixture over $t \in [T]$ and additional random noise (hence, the modified algorithm for a nonconvex loss outputs a randomized predictor). This will be apparent from the following proof.

\begin{proof}[Proof of \Cref{thm:main-al,thm:main-al-glm}]
Let $\rn{t} \in [T]$ be a random time index with $\Pr[\rn{t} = s] \propto s$. Let $q_t$ and $r_t$ denote, respectively, the queries to and responses from the conditional multiaccuracy oracle in \cref{alg:main-al}. Suppose first that $k_\ell$ is $L$-Lipschitz and set $f = k_\ell$. By \cref{thm:general-regret,thm:ftl-smooth}, with linear weights $a_t = t$, we have the following regret bound, which holds pointwise for all $x \in \X$:
\[
    \E_{\rn{t}}[\ell(r_{\rn{t}}(x), (k_\ell \circ q_{\rn{t}})(x))] \le \min_{c : \X \to [\pm 1]}\, \E_{\rn{t}}[\ell(r_{\rn{t}}(x), c(x))] + O{\Bigl(\frac{L}{T}\Bigr)}.
\]
Next, we apply indistinguishability, as described in the technical overview. Indeed, since $\ell$ is naturally parametrized, we have $\ell(r, a) - \ell(y, a) = a(y - r)$. Therefore, on the left-hand side, where the predictor is $k_\ell \circ q_t$, we may replace $r_t(\rn{x})$ with $\rn{y}$ by multiaccuracy conditional on $R_\gamma \circ k_\ell \circ q_t$ while introducing error $O(\eps + \gamma)$.\footnote{Recall that $R_\gamma$ denotes rounding to multiples of $\gamma$.} On the right-hand side, we may similarly replace $r_t(\rn{x})$ with $\rn{y}$ whenever $c \in \C$ while introducing error $O(\eps)$. Ultimately, these substitutions yield
\[
    \E_{\rn{t},\rn{x},\rn{y}}[\ell(\rn{y}, (k_\ell \circ q_{\rn{t}})(\rn{x}))] \le \min_{c \in \C}\, \E_{\rn{t},\rn{x},\rn{y}}[\ell(\rn{y}, c(\rn{x}))] + O{\Bigl(\frac{L}{T} + \gamma + \eps\Bigr)}.
\]
In the special case that $L = O(1)$, the error terms on the right-hand side are $O(\eps)$ when we set $\gamma = \eps$ and $T = 1/\eps$. Otherwise, we apply \cref{thm:loss-smoothing} to obtain an approximation $\ell_\sigma$ of $\ell$ for which $k_{\ell_\sigma}$ is $O(1/\sigma)$-Lipschitz. Letting $f = k_{\ell_{\sigma}}$ and applying the approximation guarantee from \cref{thm:loss-smoothing},
\[
    \E_{\rn{t},\rn{x},\rn{y}}[\ell(\rn{y}, (k_{\ell} \circ \rn{N}_\sigma \circ q_{\rn{t}})(\rn{x}))] \le \min_{c \in \C}\, \E_{\rn{t},\rn{x},\rn{y}}[\ell(\rn{y}, c(\rn{x}))] + O{\Bigl(\frac{1}{\sigma T} + \sigma + \gamma + \eps\Bigr)}.
\]
The error terms on the right-hand side are $O(\eps)$ when we set $\sigma = \gamma = \eps$ and $T = 1/\eps^2$.

Finally, suppose in addition that $\ell$ is convex in its second input. Then, in the smooth case, Jensen's inequality allows us to move the expectation over $\rn{t}$ as follows:
\[
    \E_{\rn{x},\rn{y}}\bigl[\ell\bigl(\rn{y}, \E_{\rn{t}}[(k_{\ell} \circ q_{\rn{t}})(\rn{x})]\bigr)\bigr] \le \min_{c \in \C}\, \E_{\rn{t},\rn{x},\rn{y}}[\ell(\rn{y}, c(\rn{x}))] + O{\Bigl(\frac{L}{T} + \gamma + \eps\Bigr)}.
\]
Similarly, in the nonsmooth case, Jensen's inequality allows us to move both the expectation over $\rn{t}$ and the expectation over the random noise:
\[
    \E_{\rn{x},\rn{y}}\bigl[\ell\bigl(\rn{y}, \E_{\rn{t}}[(k_{\ell_\sigma} \circ q_{\rn{t}})(\rn{x})]\bigr)\bigr] \le \min_{c \in \C}\, \E_{\rn{t},\rn{x},\rn{y}}[\ell(\rn{y}, c(\rn{x}))] + O{\Bigl(\frac{1}{\sigma T} + \sigma + \gamma + \eps\Bigr)}. \qedhere
\]
\end{proof}

\subsection{Achieving Omniprediction and Multicalibration} 

Finally, we show how to go beyond agnostic learning to (smooth) omniprediction and (smooth) swap multicalibration, as described in \cref{alg:main}, while still only accessing the data distribution through a multiaccuracy oracle. The proof of \cref{thm:main}, which analyzes \cref{alg:main}, follows roughly the same template as the previous proof, with a few additional steps.

\begin{proof}[Proof of \Cref{thm:main} (Omniprediction)]
    As in the preceding proof, let $\rn{t} \in [T]$ be a random time index with $\Pr[\rn{t} = s] \propto s$, let $q_t$ and $r_t$ denote the queries to and responses from the oracle, and suppose that $k_\ell$ is $L$-Lipschitz. (As usual, we also assume that $\ell$ is naturally parametrized and that $k_\ell$ is $[\pm 1]$-valued.) As in the previous proof, we apply \cref{thm:general-regret,thm:ftl-smooth} pointwise, which gives the following regret bound for linear weights $a_t = t$:
    \[
        \E_{\rn{t}}[\ell(r_{\rn{t}}(x), (k_\ell \circ q_{\rn{t}})(x))] \le \min_{c : \X \to [\pm 1]}\, \E_{\rn{t}}[\ell(r_{\rn{t}}(x), c(x))] + O{\Bigl(\frac{L}{T}\Bigr)}.
    \]

    Next, we apply indistinguishability as in the previous proof, but with some slight modifications. Once again, since $\ell$ is naturally parametrized, we have $\ell(r, a) - \ell(y, a) = a(y - r)$. Therefore, on the left-hand side, where the predictor is $k_\ell \circ q_t$, we may replace $r_t(\rn{x})$ with $\rn{y}$ by multiaccuracy conditional on the (loss-independent) quantity $R_\gamma \circ q_t$ while introducing error $O(\eps + L\gamma)$. Here, we have used the fact that $k_\ell$ is $L$-Lipschitz. On the right-hand side, we may similarly replace $r_t(\rn{x})$ with $\rn{y}$ while introducing error $O(\eps)$ whenever $c \in \C$. Ultimately, these substitutions yield
    \[
        \E_{\rn{t},\rn{x},\rn{y}}[\ell(\rn{y}, (k_\ell \circ q_{\rn{t}})(\rn{x}))] \le \min_{c \in \C}\, \E_{\rn{t},\rn{x},\rn{y}}[\ell(\rn{y}, c(\rn{x}))] + O{\Bigl(\frac{L}{T} + L\gamma + \eps\Bigr)}.
    \]
    In the special case that $L = O(1)$, the error terms on the right-hand side are $O(\eps)$ when we set $\gamma = \eps$ and $T = O(1/\eps)$. Otherwise, we apply \cref{thm:loss-smoothing} to obtain an approximation $\ell_\sigma$ of $\ell$ for which $k_{\ell_\sigma}$ is $O(1/\sigma)$-Lipschitz. Applying the approximation guarantee from \cref{thm:loss-smoothing},
    \[
        \E_{\rn{t},\rn{x},\rn{y},\rn{N}_\sigma}[\ell(\rn{y}, (k_\ell \circ \rn{N}_\sigma \circ q_{\rn{t}})(\rn{x}))] \le \min_{c \in \C}\, \E_{\rn{t},\rn{x},\rn{y}}[\ell(\rn{y}, c(\rn{x}))] + O{\Bigl(\frac{1}{\sigma T} + \frac{\gamma}{\sigma} + \sigma + \eps\Bigr)}.
    \]
    The error terms on the right-hand side are all $O(\eps)$ when we set $\sigma = \eps$ and $\gamma = \eps^2$ and $T = 1/\eps^2$. Since the loss function $\ell$ was arbitrary and no step of the algorithm depended on its choice, we conclude that \cref{alg:main} achieves omniprediction.
\end{proof}

\begin{proof}[Proof of \Cref{thm:main} (Multicalibration)]
    Departing slightly from the previous proof, let $\rn{t} \in [T]$ be uniform and let $q_t$ and $r_t$ denote the queries to and responses from the oracle with geometrically increasing weights $a_t/A_t = \lambda$ for $t \ge 2$. By \cref{thm:ftl-cal}, our online calibration error bound, we have the following inequality for all $x \in \X$ whenever the weight function $w : [\pm 1] \to [\pm 1]$ is $L$-Lipschitz:
    \[
        \bigl\lvert \E_{\rn{t}}[w(q_{\rn{t}}(x))(r_{\rn{t}}(x) - q_{\rn{t}}(x))]\bigr\rvert \le O{\Bigl(\frac{1}{\lambda T} + L \lambda \Bigr)}.
    \]
    By applying the same indistinguishability argument as in the preceding two proofs, we see that for all $c \in \C^{\mathrm{swap}}$ such that $c(\cdot, x)$ is $L$-Lipschitz for all $x \in \X$ and $c(v, \cdot) \in \C$ for all $v \in [\pm 1]$,
    \[
        \bigl\lvert \E_{\rn{t},\rn{x},\rn{y}}[c(q_{\rn{t}}(\rn{x}), \rn{x})(\rn{y} - q_{\rn{t}}(\rn{x}))]\bigr\rvert \le O{\Bigl(\frac{1}{\lambda T} + L \lambda + L \gamma + \eps \Bigr)}.
    \]
    Once again, note that there is an $O(L \gamma)$ term that arises because we round $q_t$ to multiples of $\gamma$ before calling the conditional multiaccuracy oracle.
    
    In the case of smooth swap multicalibration, where $L = O(1)$, the error terms on the right-hand side are $O(\eps)$ when we set $\lambda = \gamma = \eps$ and $T = 1/\eps^2$. For (full) swap multicalibration, we apply \cref{thm:weight-smoothing}, which uses the noise operator $\rn{N}_\sigma$ to mimic the effect of smoothing an initially nonsmooth weight function. Substituting $L = O(1/\sigma)$, the lemma implies that
    \[
        \bigl\lvert \E[c(\rn{N}_\sigma(q_{\rn{t}}(\rn{x})), \rn{x})(\rn{y} - \rn{N}_\sigma(q_{\rn{t}}(\rn{x})))]\bigr\rvert \le O{\Bigl(\frac{1}{\lambda T} + \frac{\lambda}{\sigma} + \frac{\gamma}{\sigma} + \sigma + \eps \Bigr)}.
    \]
    The error terms on the right-hand side are $O(\eps)$ when we set $\sigma = \eps$, $\lambda = \gamma = \eps^2$, and $T = 1/\eps^3$.
\end{proof}

\subsection{Deferred Lemmas on FTL and Smoothing}
\label{sec:upper-lemmas}

We now state the various helper lemmas that we used in the preceding sections. Some of these lemmas are essentially standard results in the literature with known proofs, but others are genuinely new. We defer all proofs to \cref{sec:upper-deferred}.

We begin with a helpful lemma relating the entropy function $h_\ell$ and best-response function $k_\ell$ of a given loss function $\ell$. In particular, the lemma will imply that Lipschitzness of $\partial \ell \circ k_\ell$ implies smoothness of $h_\ell$. To state the lemma, recall that for a given loss function $\ell$, we define its associated entropy to be the concave function $h_\ell(p) = \inf_a\, \ell(p, a)$ and its divergence to be the nonnegative function $D_\ell(p, a) = \ell(p, a) - h_\ell(p)$.

\begin{lemma}
\label{thm:h-vs-k}
    If $k_\ell$ is a best-response function of $\ell$, then $h'_\ell = \partial \ell \circ k_\ell$ is a superderivative of $h_\ell$ and
    \[
        D_\ell(p, k_\ell(q)) = D_{h_\ell}(p, q),
    \]
    where $D_{h_\ell}$ denotes the Bregman divergence from $p$ to $q$ with respect to $h_\ell, h'_\ell$ (\cref{def:bregman}).
\end{lemma}

Next, we present the standard regret bound for Follow the Leader. As described in the technical overview, Follow the Leader is an online algorithm which, at time $t$, performs the best action in hindsight for the $t-1$ previous time steps. The regret bound is folklore in the literature, but since it is usually stated with rather different notation, we include a self-contained proof in our context for the reader's convenience.

\begin{lemma}[Follow the Leader's Regret]
\label{thm:general-regret}
    Given a loss function $\ell$, weights $w_t \ge 0$, distribution means $p_1, \ldots, p_T \in [\pm 1]$, define the prefix sums $W_t = w_1 + \cdots + w_t$ and weighted averages $\bar{p}_t = (w_1p_1 + \cdots + w_tp_t)/W_t$ (let $\bar{p}_0 = 0$). In this setting, Follow the Leader, which in round $t$ plays the best response $k_\ell(\bar{p}_{t-1})$ in hindsight to the first $t-1$ rounds, achieves the following regret bound:
    \[
        \sum_{t=1}^T w_t\, \ell(p_t, k_\ell(\bar{p}_{t-1})) - \min_{a \in \A }\, \sum_{t=1}^T w_t\, \ell(p_t, a) = \sum_{t=1}^T W_t\, D_{\ell}(\bar{p}_t, k_\ell(\bar{p}_{t-1})).
    \]
\end{lemma}

While \cref{thm:general-regret} gives a clean exact expression for the regret, there is no reason to believe that this regret is small a priori. The following lemma, which is also more or less standard up to changes in notation, outlines an important special case where this regret can be bounded.

\begin{lemma}
\label{thm:ftl-smooth}
    In the setting of \cref{thm:general-regret}, if $\partial \ell \circ k_\ell$ is $\beta$-Lipschitz, then the regret satisfies
    \[
        \sum_{t=1}^T W_t D_\ell(\bar{p}_t, k_\ell(\bar{p}_{t-1})) \le O{\Bigl(\beta\sum_{t=1}^T \frac{w_t^2}{W_t}\Bigr)}.
    \]
\end{lemma}

The preceding lemma bounded the regret of Follow the Leader in terms of a sum $\sum_t w_t^2/W_t$ depending on the chosen weights $w_t$. The next technical lemma shows that this bound is indeed small for many choices of the weights, such as uniform weights $w_t = 1$, linearly increasing weights $w_t = t$, and geometrically increasing weights $w_t$ with $w_t/W_t = \lambda$ for all $t \ge 2$. In each case, we normalize the weights so that they sum to $W_T = 1$. To our knowledge, the lemma is new, but nevertheless quite straightforward.

\begin{lemma}[Step Sizes]
\label{thm:step-sizes}
    Given $w_t \ge 0$, let $W_t = w_1 + \cdots + w_t$ and $\delta = \sum_{t=1}^T w_t^2/W_t W_T$. Then,
    \begin{itemize}
        \item \textbf{(Uniform)} If $w_t = 1$ for all $t \ge 1$, then $\delta = O(\log(T)/T)$.
        \item \textbf{(Linear)} If $w_t = t$ for all $t \ge 1$, then $\delta = O(1/T)$.
        \item \textbf{(Geometric)} If $w_t/W_t = \lambda = \log(T)/T$ for $t \ge 2$, then $\delta = \lambda + (1-\lambda)^T = O(\log (T)/T)$.
    \end{itemize}
\end{lemma}

Now, we shift our attention to the online smooth calibration error of Follow the Leader, as opposed to its regret for a given loss function. As discussed, deterministic online algorithms for smooth calibration were previously known, but to our knowledge, the fact that such a simple algorithm as Follow the Leader achieves this goal was not previously known. The following lemma uses geometrically increasing weights.

\begin{lemma}[Follow the Leader's Calibration Error]
\label{thm:ftl-cal}
    Given a $\beta$-Lipschitz function $w : [\pm 1] \to [\pm 1]$ and distribution means $p_1, \ldots, p_T \in [\pm 1]$, define the weighted Follow the Leader iterates $\bar{p}_0 = 0$, $\bar{p}_1 = p_1$, $0 < \lambda \leq 1$, and $\bar{p}_t = (1-\lambda)\bar{p}_{t-1} + \lambda p_t$ for $t = 2, \ldots, T$. Then, we have the following online calibration bound:
    \[
        \Bigl\lvert\sum_{t=1}^T w(\bar{p}_{t-1})(p_t - \bar{p}_{t-1})\Bigr\rvert \le O{\Bigl(\frac{1}{\lambda} + \beta \lambda T\Bigr)}.
    \]
\end{lemma}

Next, we prove some useful smoothing lemmas, which will allow us to transfer results for smooth omniprediction to full omniprediction, and similarly transfer results for smooth multicalibration to full multicalibration. This is relevant, in particular, for $\ell_1$ loss's non-Lipschitz best response, needed for our main result on agnostic learning.

For loss minimization, the smoothing operation that we study can be viewed in two equivalent ways: either as a Lipschitz approximation to the best-response function $k_\ell$ or as a small addition of noise to the prediction itself. While the former view has the advantage of being deterministic, the second is more convenient for omniprediction since it is not specific to the loss $\ell$ under consideration.

\begin{lemma}[Loss Smoothing]
\label{thm:loss-smoothing}
    Let $\ell$ be a naturally parametrized loss function with a $[\pm B]$-valued best-response function $k_\ell$. Then, for all $\sigma^{-1} \in \N$, there exists a naturally parametrized loss function $\ell_\sigma$ with the following two properties:
    \begin{itemize}
        \item \textbf{(Smoothness)} The best response function $k_{\ell_\sigma}$ is $[\pm B]$-valued and $O(B/\sigma)$-Lipschitz.
        \item \textbf{(Approximation)} For all $p\in[\pm1]$ and $\eta \in \R$,
        \[
            \ell_\sigma(p, \eta) \le \ell(p, \eta) + O(B\sigma).
        \]
        Also, for all $p,q \in [\pm 1]$,
        \[
            \bigl\lvert \E[\ell(p, k_{\ell}(\rn{N}_\sigma(q)))] - \ell_\sigma(p, k_{\ell_\sigma}(q))\bigr\rvert \le O(B\sigma).
        \]
    \end{itemize}
    Concretely, $\ell_\sigma$ and $k_{\ell_\sigma}$ are derived from $\ell$ and $k_\ell$ by the formula
    \[
        k_{\ell_\sigma}(p) = \E[k_\ell(\rn{N}_\sigma(p))].
    \]
    Here, $\rn{N}_\sigma : [\pm 1] \to [\pm 1]$ is the (randomized) noise function that randomly rounds its input either up or down to the nearest multiple of $\sigma$ while preserving its input's value in expectation.
\end{lemma}

For example, starting from the $\ell_1$ loss with best response function $\sign(p)$ the smoothing lemma constructs a modified loss whose best response function is $\E[\sign(\rn{N}_\sigma(p))]$. If we use the convention $\sign(0) = 0$, then this is precisely the piecewise linear function $\sign_\sigma(p)$ defined in the introduction, which agrees with $\sign(p)$ when $|p| \ge \sigma$ and increases linearly from $-1$ to $+1$ over the interval $[\pm \sigma]$.

Finally, we state our lemma for smoothing calibration weight functions by adding random noise. This lemma is essentially known in the recent literature \cite{blasioksmooth,HartlineWY25,gopalan2026smce}, but for slightly different noise functions than the one we consider. Our chosen noise function, which we denote by $\rn{N}_\sigma$, has the advantage of keeping the predictions discrete-valued in a grid over the interval $[\pm 1]$.

\begin{lemma}[Weight Smoothing]
\label{thm:weight-smoothing}
    Suppose that $p$ has calibration error $\eps_{L}$ for weight functions $c \in \C^{\mathrm{swap}}$ that are $[\pm 1]$-valued and $L$-Lipschitz in their first input, meaning that
    \[
        \E[c(p(\rn{x}), \rn{x})(\rn{y} - p(\rn{x}))] \le \eps_L.
    \]
    Then, $\rn{N}_\sigma \circ p$ has calibration error $\eps_{O(1/\sigma)} + O(\sigma)$ for arbitrary $[\pm 1]$-valued weights $c \in \C^{\mathrm{swap}}$.
\end{lemma}

\section{Lower Bounds}
\label{sec:lower}

In this section, we give examples showing that the number of calls that we make to the conditional multiaccuracy oracle is essentially tight in several important cases. We emphasize that our lower bounds are on the number of calls to a \emph{conditional} multiaccuracy oracle. Proving lower bounds on the required number of calls to the underlying multiaccuracy oracle, as opposed to the conditional multiaccuracy gadget, is an interesting direction for future work.

\subsection{Lower Bounds for Omniprediction}

We start by showing via the guiding example from \cref{sec:symmetry-example} that the $O(1/\eps)$ upper bound for smooth omniprediction and the $O(1/\eps^2)$ upper bound for (full) omniprediction are both tight up to constant factors. As our argument will show, they are tight not only for our specific algorithm, but also for any algorithm whose queries and final output are restricted to be pointwise functions of the oracle responses of a conditional multiaccuracy oracle. In fact, this example rules out improvements even for $\ell_2$ and $\ell_1$ loss minimization, which are clearly prerequisites for smooth omniprediction and (full) omniprediction, respectively.

\begin{theorem}[Large $\ell_1, \ell_2$]
\label{thm:l1-l2-lower}
    Consider the class of algorithms with access to a conditional multiaccuracy oracle, whose queries and final output are restricted to be pointwise functions of the oracle responses received so far. Among such algorithms, $\Omega(1/\eps)$ queries are required to achieve $\eps$-approximate $\ell_2$ loss minimization, and $\Omega(1/\eps^2)$ queries instead for $\ell_1$ loss.
\end{theorem}

\begin{proof}
Recall the guiding example in \cref{sec:symmetry-example}, where the data distribution consists of uniform $n$-bit features $\rn{x} \sim \{\pm 1\}^n$ for odd $n$, and binary labels $\rn{y} = \rn{x}_n$, while $\C$ is the class of \emph{symmetric} functions $c : \{\pm 1\}^n \to [\pm 1]$, which are invariant under permutations of the input bits. The predictors $p_i(\x) = \x_i$ for $i \leq n-1$ are each independent of $\y$, and moreover the same is true for any post-processing function $f : [\pm 1]^{n-1} \to [\pm 1]$ used to combine these $n-1$ predictors:
\[\E[\x_nf(\rn{x}_1, \ldots, \rn{x}_{n-1})] = 0.\]

The $\ell_1$ loss of any function $f$ of $p_1,\ldots, p_{n-1}$ is $1$ since
\[
	\E\bigl[\lvert \rn{x}_n - f(\rn{x}_1, \ldots, \rn{x}_{n-1}) \rvert \bigr] = 1 - \E[\x_nf(\rn{x}_1, \ldots, \rn{x}_{n-1})] = 1.
\]

On the other hand, the Majority function $\mathrm{Maj}_n$, which is clearly symmetric, achieves correlation $\Omega(1/\sqrt{n})$.\footnote{Since we assume $n$ is odd, $\mathrm{Maj}_n$ is well-defined.} Using the standard fact that its degree-$1$ Fourier coefficients are all $\Omega(1/\sqrt{n})$,
\[
    \E\bigl[|\rn{x}_n - \mathrm{Maj}_n(\rn{x})|\bigr] = 1 - \Omega{\Bigl(\frac{1}{\sqrt{n}}\Bigr)}.
\]
Unless $n > \Omega(1/\eps^2)$, the $\Omega(1/\sqrt{n})$ gap between the loss suffered by $f(x_1,\ldots, x_{n-1})$ and $\mathrm{Maj}_n$ exceeds $\eps$.

We now consider the $\ell_2$ loss, which can be bounded for any $f:\{\pm 1\}^{n-1} \to [\pm 1]$ as
\begin{align*}
	\E\bigl[(\x_n - f(\x_1,\ldots,\x_{n-1}) )^2 \bigr] &= 1 - 2\E[\x_n f(\x_1,\ldots,\x_{n-1})] + \E[f(\x_1,\ldots, \x_{n-1})^2] \geq 1. 
\end{align*}

The optimal symmetric function is $s_n(\x) = \sum_i \x_i/n$, which is perfectly calibrated and whose sign is Majority.
Computing its loss, we see that
\[
    \min_{c \in \C}\, \E\bigl[(\rn{y} - c(\rn{x}))^2\bigr] \le \E\biggl[\Bigl(\rn{x}_n - \frac{\rn{x}_1 + \cdots + \rn{x}_n}{n}\Bigr)^2\biggr] = 1 - \frac{1}{n}.
\]
Unless $n > \Omega(1/\eps)$, the gap between the $\ell_2$ loss suffered by $f$ and $s_n$ exceeds $\eps$.
\end{proof}

\subsection{Lower Bounds for Multicalibration}

What remains is to show that our $O(1/\eps^2)$ upper bound for smooth swap multicalibration and $O(1/\eps^3)$ upper bound for (full) swap multicalibration are also tight up to constant factors. In fact, our examples will rule out improvements even for smooth calibration and (full) calibration, where there is effectively no class $\C$. As a bonus, we will also show that the randomization over iterates performed at the end of \cref{alg:main} is necessary, in the sense that there may not exist any particular iterate, besides the initial $0$ predictor, that is smoothly calibrated. We emphasize that the following lower bounds, unlike \cref{thm:l1-l2-lower}, are quite specific to our specific oracle algorithm, \cref{alg:main}.

To state our various results, it will be helpful to define the notion of a \emph{valid} sequence of queries $q_0, \ldots, q_T$ to a conditional multiaccuracy oracle.\footnote{In the previous section, the first query was $q_1$, but in this section, it will be cleaner to use zero-based indexing of the queries $q_0, q_1, \ldots$ with the initialization $q_0 = 0$.} These are query sequences that could have arisen over the course of \cref{alg:main} for some underlying data distribution and some adversarially chosen sequence of oracle responses. The hard problem instances that we construct do not assume the class $\C$ contains any functions besides the constant functions, nor do they assume that any error, such as  statistical or rounding error, is present in the oracle responses. We will, however, place a restriction on the types of queries submitted to the oracle; we require these to be weighted averages of the previous responses with coefficients determined in advance by a constant step size $\lambda$. This restriction was not needed for \cref{thm:l1-l2-lower}.

\begin{definition}[Valid Query Sequence]
\label{def:valid-sequence}
    A sequence of query predictors $q_1, \ldots, q_T : \X \to [\pm 1]$ is \emph{valid for $\lambda \in (0, 1]$} if there exists a distribution $\D$ over pairs $(\rn{x},\rn{y}) \in \X \times \{\pm1\}$ and a sequence of response predictors $r_1, \ldots, r_T: \X \to [\pm 1]$ such that the following conditions hold for all $t=1, \ldots, T$:
    \begin{itemize}
        \item (Update Rule) if $\lambda_1 = 1$ and $\lambda_t = \lambda$ for $t \ge 2$, then
        \[
            q_t = (1-\lambda_t)q_{t-1} + \lambda_t r_t;
        \]
        \item (Conditional Accuracy in Expectation)
        \[
            \E[\rn{y} - r_t(\rn{x})\mid q_{t-1}(\rn{x})] = 0.
        \]
    \end{itemize}
    We use the convention $q_0 = 0$.
\end{definition}

Our first two results of this section, \cref{thm:smce-lower,thm:ece-lower}, state that there exist long valid sequences with large calibration error. By ``large,'' we mean greater than $\eps$, and by ``long,'' we mean $T = \Omega(1/\eps^2)$ for degree-2 calibration or $T = \Omega(1/\eps^3)$ for (full) calibration. These lemmas show that our $T = O(1/\eps^2)$ upper bound for smooth swap multicalibration and $T = O(1/\eps^3)$ upper bound for (full) swap multicalibration are tight. We defer all further proofs in this section to \cref{sec:lower-deferred}.

\begin{lemma}[Large $\TwoCE$]
\label{thm:smce-lower}
    For all $\lambda \in (0, 1]$ and $T \in \N$, there exist valid $q_1, \ldots, q_T$ for $\lambda$ such that
    \[
        \min_{t \in [T]}\,|\TwoCE(q_t)| = \Omega{\Bigl(\frac{1}{\sqrt{T}}\Bigr)}.
    \]
    In fact, we have the stronger lower bound $\Omega(\min\{1, 1/(\lambda T) + \lambda\})$, which also holds for the degree-2 calibration error of any randomized predictor defined as a random mixture of $q_1, \ldots, q_T$ composed with additional unbiased random noise $\rn{N} : [\pm 1] \to [\pm 1]$.
\end{lemma}

\begin{lemma}[Large $\ECE$]
\label{thm:ece-lower}
    For all $\lambda \in (0, 1]$ and $T \in \N$, there exist valid $q_1, \ldots, q_T$ for $\lambda$ such that 
    \[
        \min_{t \in [T]}\, \ECE(q_t) = \Omega{\Bigl(\frac{1}{\sqrt[3]{T}}\Bigr)}.
    \]
    In fact, we have the stronger lower bound $\Omega(\min\{1, 1/(\lambda T) + \sqrt{\lambda}\})$, which also holds for the expected calibration error of any randomized predictor defined as a random mixture of $q_1, \ldots, q_T$ composed with additional unbiased random noise $\rn{N} : [\pm 1] \to [\pm 1]$.
\end{lemma}

Our next result, \cref{thm:smce-mixture}, shows that the random mixture over time steps at the end of \cref{alg:main} is necessary, in the sense that one cannot hope to simply output a single, deterministic iterate $q_t$ from the sequence. Crucially, the hard instances that we construct apply when there are multiple weight functions of interest. This is not a coincidence. As we saw in \cref{sec:upper} and will discuss further in \cref{sec:deg-2-derand}, when there is only one weight function or loss function of interest (e.g. agnostic learning) \cref{alg:main} can indeed be modified to output a single iterate $q_t$, possibly with additional noise and post-processing.

\begin{lemma}[Smooth Calibration Tradeoff]
\label{thm:smce-mixture}
    For all $\lambda \in (0, 1]$, there exist valid $q_1, q_2, \ldots$ such that
    \[
       \min_{t \in \N} \max_{i \in [2]}\, |\CE_{w_i}(q_t)| = \Omega(1)
    \]
    for two different $1$-Lipschitz and $[\pm1]$-valued weight functions $w_1,w_2 : [\pm 1] \to [\pm 1]$.
\end{lemma}

Although we defer the proofs of \cref{thm:smce-lower,thm:ece-lower,thm:smce-mixture} to \cref{sec:lower-deferred}, we sketch the main ideas here.

The results  all rely on a certain simple family of predictors, which we will use for the Bayes-optimal $p^*$, the queries $q_t$, and the responses $r_t$. In this family, the feature space is $\X = \sqrt{n}\cdot S^{n-1}$, the sphere of radius $\sqrt{n}$ centered at $0 \in \R^n$, and each $a \in \R^n$ defines a predictor
\[
    a(x) = \langle a, x\rangle.
\]
This predictor is $[\pm 1]$-valued as long as $\|a\| \le 1/\sqrt{n}$. For clarity of presentation, our examples will sometimes involve vectors $p^*, q_t, r_t \in \R^n$ with norm greater than $1/\sqrt{n}$, but still $O(1)$. We keep in mind that these vectors must eventually be scaled down to have norm at most $1/\sqrt{n}$, but this scaling will not affect any of our results since all our examples use $n = 2$. Intuitively, the sequences $q_t$ that we construct will resemble circular or oscillating paths in the plane.

There are a few key properties of these predictors that we use repeatedly. They pertain to expectations of products, conditional expectations, and expected absolute values. We state the properties for arbitrary dimensions $n \in \N$, but to reiterate, we only really need the case $n = 2$.

\begin{lemma}
\label{thm:geom}
    Fix $n \ge 2$ and sample $\rn{x} \sim \Unif(\sqrt{n}\cdot S^{n-1})$. Then for all $a,b \in \R^n$ with $\|b\|=1$,
    \begin{align*}
        \E\bigl[\langle a, \rn{x}\rangle \langle b, \rn{x}\rangle\bigr] &= \langle a, b \rangle,\\
        \E\bigl[\langle a, \rn{x}\rangle \mid \langle b, \rn{x}\rangle\bigr] &= \langle a, b \rangle \langle b, \rn{x} \rangle.
    \end{align*}
    In particular, $\E[\langle a, \rn{x}\rangle \mid \langle b, \rn{x} \rangle] = 0$ if and only if $\langle a, b \rangle = 0$. Also,
    \[
        \E\bigl[|\langle b, \rn{x} \rangle|\bigr] = C_n,
    \]
    where $3/4 \le C_n \le 1$ is a constant depending only on $n$, and not the direction of $b$.
\end{lemma}

The first key observation is that in the linear predictor family, degree-2 calibration error $\TwoCE$ measures the squared distance to the point $p^*/2$. Indeed, by \cref{thm:geom}, for any $p^*, q \in \R^n$,
\begin{equation}
\label{eq:twoce-formula}
    \TwoCE(q) = \E[q(\rn{x})(\rn{y} - q(\rn{x}))] = \langle q, p^* - q \rangle = \Bigl\|\frac{p^*}{2}\Bigr\|^2 - \Bigl\|q - \frac{p^*}{2}\Bigr\|^2.
\end{equation}
In other words, if $n = 2$ and a sequence of points $q_1, \ldots, q_T$ lies entirely on a circle centered at the point $p^*/2$, then the sequence has constant degree-2 calibration error (in $n \ge 3$ dimensions, this set becomes a sphere). For a non-zero prediction vector, \cref{thm:geom} implies a similar formula for expected calibration error:
\begin{equation}
\label{eq:ece-formula}
    \ECE(q) = \E\bigl[\lvert\E[\rn{y}|q(\rn{x})] - q(\rn{x})\rvert\bigr] = \E\biggl[\Bigl\lvert \frac{\langle p^*, q \rangle}{\|q\|^2}\langle q, \rn{x} \rangle - \langle q, \rn{x}\rangle\Bigr\rvert\biggr]
    = C_n \frac{\lvert\langle q, p^* - q \rangle\rvert}{\|q\|}.
\end{equation}
Here, $C_n \in [3/4, 1]$ is the constant from \cref{thm:geom}. In particular, for the family of predictors under consideration, degree-2 and expected calibration error essentially differ by a factor of $\|q\|$:
\[
    \ECE(q) = C_n \frac{|\TwoCE(q)|}{\|q\|}.
\]
For the zero predictor, both calibration errors are zero.
We also have the general relationship $\ECE(q) \ge \lvert \TwoCE(q) \rvert$, which is not specific to the linear families under consideration. For the remaining details of the constructions, see \cref{sec:lower-deferred}.

\section{Derandomization}
\label{sec:derand}

In several cases, our learning algorithms output \emph{randomized} predictors $\rn{p} : \X \to [\pm 1]$, which may output different values in $[\pm 1]$ on the same input $x \in \X$ when invoked multiple times. In practice, this trait is arguably undesirable, raising the question of whether these predictors can be derandomized while retaining their various learning guarantees. In this section, we study several cases where this derandomization can indeed be performed.

We already saw that our algorithm does not require randomness to achieve loss minimization with respect to any single loss function $\ell$ with Lipschitz $k_\ell$. Next, in \cref{sec:deg-2-derand}, we similarly show that our algorithm can achieve calibration with respect to any single Lipschitz weight function without introducing unnecessary randomness. To this end, we first use the weight function $w(p) = p$ (i.e., degree-$2$ calibration $\TwoCE$) as a case study, to convey the intuition behind the proof, and then state the derandomization result for the more general case of $L$-Lipschitz weight functions $w$.
Next, in \cref{sec:general-derand}, we give a recipe for derandomizing a predictor in a black-box fashion while maintaining a wide range of other learning guarantees.

\subsection{Degree-2 Calibration}
\label{sec:deg-2-derand}

Suppose throughout that the query-response pairs $(q_t,r_t)$ satisfy $\mathbb{E}[\rn{y}-r_{t+1}(\x)\mid q_t(\x)]=0$ (i.e., they satisfy the conditional accuracy in expectation condition from Definition~\ref{def:valid-sequence}).
Recall that the geometric update is defined as
\[
    q_{t+1}
    =
    (1-\lambda)q_t+\lambda r_{t+1}
    =
    q_t+\lambda(r_{t+1}-q_t),
\]
where $q_t,r_t:\mathcal{X}\to[\pm1]$. In this subsection only, we ignore oracle approximation and prediction-rounding errors for simplicity, and work with exact conditional accuracy.

At each step $t$, we denote the degree-2 calibration error by $\alpha_t := \TwoCE(q_t) = \E[q_t(\x)(\y-q_t(\x))]$. 
Our goal is to show that every sufficiently long consecutive sequence of query predictors $\{q_t\}$ contains at least one (deterministic) predictor whose absolute degree-2 calibration error satisfies $|\alpha_t| \leq \alpha$, and hence is degree-2 calibrated with error parameter $\alpha$.

The proof proceeds in two steps:
\begin{enumerate}
    \item \textbf{Consecutive violations are close (Claim~\ref{claim:alpha_t-dist}).}
    Per the geometric update, each successive update changes the query predictor $q_t$ by at most $2\lambda$, i.e., $|q_{t+1}(x)-q_t(x)| \leq 2 \lambda$.
    Since the degree-2 calibration violation is Lipschitz in the prediction (specifically, the map $v \mapsto v(y-v)$ is 3-Lipschitz),
    we can translate the closeness between successive $q_t$ updates to closeness between successive $\alpha_t$ errors.
    This means that consecutive violations cannot switch from substantially positive to substantially negative (or vice versa) without passing through the approximately calibrated region $[-\alpha, \alpha]$.
    \item \textbf{Telescoping bounds the average signed violation (Theorem~\ref{thm:degree-2-cal}).}
    Summing the identity in Lemma~\ref{lemma:potential-q_t} over $N$ consecutive rounds and telescoping gives
    \[
        \Bigg| \dfrac{1}{N} \sum_t \alpha_t \Bigg| \leq 
        O \Big(\dfrac{1}{\lambda N} + \lambda\Big).
    \]
    Under the step size $\lambda$ and sequence length assumption in Theorem~\ref{thm:degree-2-cal}, the RHS is upper-bounded by $\alpha$.
    This alone does not imply that any one $|\alpha_t|$ violation is small. 
    However, Claim~\ref{claim:alpha_t-dist} is what excludes this possibility.
    Indeed, suppose that $|\alpha_t| > \alpha$ for every round in these $N$ consecutive rounds.
    By Claim~\ref{claim:alpha_t-dist}, consecutive violations cannot jump across the approximately calibrated region.
    Since none lies inside that region, then they must all either satisfy $\alpha_t > \alpha$ or all satisfy $\alpha_t < -\alpha$.
    Their average would likewise also be greater than $\alpha$ or less than $-\alpha$, respectively.
    In either case, 
    \[
        \Bigg|  \dfrac{1}{N} \sum_t \alpha_t \Bigg| > \alpha,
    \]
    which contradicts the initial telescoping inequality.
\end{enumerate}

Note that it is not necessarily true that the absolute violation $|\alpha_t|$ decreases at every round.
What we show is that the query predictors cannot all have large absolute violation for too many consecutive rounds.

The first step of the proof can be formally stated as follows:
\begin{claim}\label{claim:alpha_t-dist}
    For all $t \geq 1$, 
    $|\alpha_{t+1}-\alpha_t| \leq 6\lambda$. 
\end{claim}

The proof of this claim follows from Lemma 7.4 in \cite{blasiok2023unifying}, adapted to the $[\pm 1]$ range of our predictors. 
The proof requires expanding the $q_{t+1}(x)(y-q_{t+1}(x)) - q_t(x)(y-q_t(x))$ expression point-wise, factoring out the $q_{t+1}-q_t$ term, and using the fact that all of $y, q_t, q_{t+1}, r_{t+1}$ output values in $[\pm 1]$.

Next, we use a potential-based argument to show that in any long-enough sequence of iterates $\{q_t\}$, at least one of the predictors is $\alpha$-degree-2 calibrated.
Formally:
\begin{theorem}\label{thm:degree-2-cal}
    Fix $\alpha>0$ and assume $\lambda \leq \alpha/4$.
    Suppose that $(q_t, r_t)$ are a valid sequence of query-response pairs that satisfy the geometric update rule. 
    Let $s \geq 1$ be any starting round. 
    If $N > 1/(\lambda \alpha)$, then there exists $t \in \{s, \ldots, s+N-1\}$ such that
    \[
        |\TwoCE(q_t)| = \bigl| \E[q_t(\x)(\y-q_t(\x))] \bigr| \leq \alpha.
    \]
    Equivalently, no consecutive block of more than
    $1/(\lambda \alpha)$ iterates can remain entirely outside the
    $\alpha$-degree-$2$ calibration region.
\end{theorem}

To prove this theorem, we use an appropriate potential function, defined using the $q_t$ values.
Before proving Theorem~\ref{thm:degree-2-cal}, we define and prove a key identity of this potential function:
\begin{lemma}\label{lemma:potential-q_t}
    Let $\Phi_t = \E[q_t(\x)^2]$. Then,
    \[
        \Phi_{t+1} - \Phi_t = 2 \lambda \alpha_t + \lambda^2 \E[(r_{t+1}(\x)-q_t(\x))^2],
    \]
    where $\lambda$ is defined in the geometric update and $\alpha_t = \E[q_t(\x)(\y-q_t(\x))]$.
\end{lemma}

\begin{proof}
    By the query-response pairs property, it follows that
    \begin{align}
        \E[q_t(\x)(\y-r_{t+1}(\x))] &= \E[q_t(\x) \cdot \E\bigl[\y-r_{t+1}(\x) | q_t(\x)]\bigr] \nonumber \\
        &= 0,\label{eq:residual-0}
    \end{align}
    since on each level set of $q_t$ (on which $q_t$ is, by definition, constant), the residual $\y-r_{t+1}$ has mean equal to 0. 
    Therefore, $\alpha_t = \E[q_t(\x)(r_{t+1}(\x)-q_t(\x))]$.
    By adding and subtracting $r_{t+1}$, we have:
    \begin{align*}
        \alpha_t &= \E[q_t(\x)\cdot (\y-r_{t+1}(\x)+r_{t+1}(\x)-q_t(\x))] \\
        &= \E[q_t(\x) \cdot (\y-r_{t+1}(\x))] + \E[q_t(\x) \cdot (r_{t+1}(\x) - q_t(\x))] \\
        &= 0+ \E[q_t(\x) \cdot (r_{t+1}(\x) - q_t(\x))] \qquad \qquad \qquad \textrm{ by equation~\eqref{eq:residual-0}}
    \end{align*}
    Pointwise, by expanding and applying the geometric update definition, we have that
    \begin{align*}
        q_{t+1}(x)^2 - q_t(x)^2 &= \bigl(q_t(x) + \lambda (r_{t+1}(x) - q_t(x))\bigr)^2 - q_t(x)^2 \\
        &= 2\lambda q_t(x) \cdot (r_{t+1}(x) - q_t(x)) + \lambda^2 \cdot (r_{t+1}(x)-q_t(x))^2.
    \end{align*}
    By taking expectations and using the fact that $\alpha_t = \E[q_t(\x)(r_{t+1}(\x)-q_t(\x))]$, it follows that
    \[
        \Phi_{t+1} - \Phi_t = 2 \lambda \alpha_t + \lambda^2 \E[(r_{t+1}(\x)-q_t(\x))^2], 
    \]
    as we wanted to show.
\end{proof}

Summing the identity in Lemma~\ref{lemma:potential-q_t} over $N$ consecutive rounds and re-arranging gives
\[
    \dfrac{1}{N} \sum_{t=s}^{s+N-1} \alpha_t =
    \dfrac{\Phi_{s+N} - \Phi_s}{2 \lambda N} - \dfrac{\lambda}{2N} \sum_{t=s}^{s+N-1} \E[(r_{t+1}(x)-q_t(x))^2].
\]
Since the potential takes values between zero and one, and each squared difference is at most four, taking absolute values yields
\[
    \Bigg| \dfrac{1}{N} \sum_t \alpha_t \Bigg| \leq \dfrac{1}{2\lambda N} + 2\lambda,
\]
which is the expression we used in the proof intuition at the beginning of Section~\ref{sec:deg-2-derand}.

With this lemma in hand, we now prove Theorem~\ref{thm:degree-2-cal}:
\begin{proof}[Proof of Theorem~\ref{thm:degree-2-cal}]
    Suppose toward contradiction that
    \begin{equation}
        |\alpha_t|>\alpha \qquad \text{for every }t\in\{s,\ldots,s+N-1\}.
        \label{eq:degree-two-entire-block-bad}
    \end{equation}
        By Claim~\ref{claim:alpha_t-dist}, \[
        |\alpha_{t+1}-\alpha_t|
        \leq
        6\lambda
        \leq
        2\alpha.
    \]
    Therefore, the values in
    equation~\eqref{eq:degree-two-entire-block-bad}
    cannot change sign. Indeed, moving from a value greater than
    $\alpha$ to a value smaller than $-\alpha$ would require a change
    strictly larger than $2\alpha$.
    Consequently, either
    \begin{equation}
        \alpha_t>\alpha
        \qquad
        \text{for every }
        t\in\{s,\ldots,s+N-1\},
        \label{eq:degree-two-positive-block}
    \end{equation}
    or
    \begin{equation}
        \alpha_t<-\alpha
        \qquad
        \text{for every }
        t\in\{s,\ldots,s+N-1\}.
        \label{eq:degree-two-negative-block}
    \end{equation}
    
    Suppose first that
    equation~\eqref{eq:degree-two-positive-block}
    holds.
    By Lemma~\ref{lemma:potential-q_t}, $ \Phi_{t+1}-\Phi_t
        >2\lambda\alpha.$
    Summing over $t=s,\ldots,s+N-1$ gives $\Phi_{s+N}-\Phi_s
        >2N\lambda\alpha.$
    Given that by assumption $N > 1/(\lambda \alpha)$, it follows that the right-hand side is greater than 2.
    This is impossible because
    $\Phi_s,\Phi_{s+N}\in[0,1]$, and hence $\Phi_{s+N}-\Phi_s\leq 1$.

    Suppose instead that equation~\eqref{eq:degree-two-negative-block}
    holds. Since $(r_{t+1}(x)-q_t(x))^2\leq4$, Lemma~\ref{lemma:potential-q_t} implies that
    \[
    \begin{aligned}
    \Phi_{t+1}-\Phi_t
    &<
    -2\lambda\alpha+4\lambda^2
    \\
    &\leq
    -\lambda\alpha,
    \end{aligned}
    \]
    since $\lambda\leq\alpha/4$.
    Summing over $t=s,\ldots,s+N-1$ gives $\Phi_{s+N}-\Phi_s
        < -N\lambda\alpha.$
    We again reach a contradiction, since $N>1/(\lambda\alpha)$ implies that the left-hand side is smaller than $-1$, which is impossible since $\Phi_s,\Phi_{s+N}\in[0,1]$, and hence $\Phi_{s+N}-\Phi_s\geq-1.$

    Hence equation~\eqref{eq:degree-two-entire-block-bad}
    cannot hold, proving that at least one of the $q_t$ predictors for some
    $t\in\{s,\ldots,s+N-1\}$ is $\alpha$-degree-2 calibrated.
\end{proof}

\subsubsection{Generalization to Fixed $w$-Calibration}\label{sec:L-lipschitz-cal}
Beyond 2-calibration error, for a weight function $w:[\pm1]\to[\pm1]$, define the $w$-calibration
violation of the query predictor at round $t$ by
\[
    \alpha_t := \CE_w(q_t)
    =
    \mathbb{E}[w(q_t(\x))(\y-q_t(\x))].
\]
In particular, taking $w(p)=p$ recovers the degree-$2$ calibration definition.

We generalize our derandomization result of the degree-2 calibration case (Theorem~\ref{thm:degree-2-cal}) to a general $L$-Lipschitz weight function $w$:
\begin{theorem}\label{thm:degree-w-cal}
    Let $s \geq 1$ be any starting round and let $w:[\pm1] \to [\pm 1]$ be $L$-Lipschitz. 
    Suppose that $(q_t, r_t)$ are a valid sequence of query-response pairs that satisfy the geometric update rule. 
    Fix $\alpha>0$ and assume $\lambda
    \leq
    \alpha/(4L+2)$ and $(T-1)(\lambda\alpha/2)>2$.
    Then there exists a fixed
    $t\in\{s,\ldots,s+T-1\}$ such that
    \[
        |\CE_w(q_t)| = 
        \left|
            \mathbb{E}[w(q_t(\x))(\y-q_t(\x))]
        \right|
        \leq
        \alpha.
    \]
\end{theorem}

The proof is very similar to the case of degree-2 calibration: we use the same geometric update and exact conditional-accuracy assumption as in the degree-2 case.
For this reason, we defer the formal proofs to the appendix.

The proof of Theorem~\ref{thm:degree-w-cal} then follows the same two steps:
\begin{enumerate}
    \item \textbf{Consecutive violations are close (Lemma~\ref{lemma:L-lips}).}
    The geometric update is unchanged, so consecutive query predictions still differ by at most $2\lambda$.
    What changes is the Lipschitz constant: the map $v \mapsto w(v)(y-v)$ is now Lipschitz with constant $2L+1$.
    Consequently, $|\alpha_{t+1}-\alpha_t| \leq 2(2L+1)\lambda$.
    Hence if $\lambda$ is small enough relative to $\alpha$, consecutive violations cannot jump across the approximately calibrated region $[-\alpha, \alpha]$.
    \item \textbf{Telescoping bounds the average signed violation (Theorem~\ref{thm:degree-w-cal}).}
    The same argument as with the degree-2 calibration case holds, this time with the telescoping expression
    \[
        \left|\frac{1}{T-1}\sum_t \alpha_t\right|
    \leq O \Big( \frac{1}{\lambda(T-1)} + L\lambda \Big).
    \]
    Under the step size $\lambda$ and sequence length assumption in Theorem~\ref{thm:degree-w-cal}, the RHS is upper-bounded by $\alpha$. 
\end{enumerate}

\subsection{General Derandomization}
\label{sec:general-derand}

We will now provide a general and simple recipe for derandomizing a randomized predictor. The main idea is to partition the space into sets of sufficiently small mass and fix, for each piece, an independent predictor drawn from the original distribution over predictors. Similar ideas have appeared in prior work in different contexts (see, e.g., \cite{lange2026robust,noarov2026deterministic}), but we provide all the relevant statements and proofs here for completeness. The proof is based on a simple Hoeffding bound.

The proposed derandomization procedure requires the existence of a partition of the feature space into low-mass regions. This assumption is captured by the following definition.

\begin{definition}[Partition Function]\label{def:partition-function}
    Let $\D_\X$ be a distribution over $\X$ and let $f:\X\to [V]$ be measurable and total, where $V\in \N$. We say that $f$ is a $\gamma$-partition function with respect to $\D_\X$ if the following inequality holds for all $v \in [V]$:
    \[
        \Pr_{\rn{x}\sim\D_\X}[f(\rn{x})=v] \le \gamma.
    \]
\end{definition}

Given a sufficiently refined partition function, we use the following simple, black-box derandomization procedure.

\begin{definition}[Piecewise Derandomization]
    Let $f$ be a $\gamma$-partition function with respect to a distribution $\D_\X$ over $\X$, and let $\rn{p}$ be a randomized predictor, i.e., a random variable over the space of functions mapping $\X$ to $[\pm 1]$. The \emph{$f$-piecewise derandomization} of $\rn{p}$ is the following randomized procedure that outputs a deterministic predictor ${p}^f$.
    \begin{itemize}
        \item For each $v\in [V]$, let $\rn{p}_v$ be an independent copy of $\rn{p}$.
        \item Let ${p}^f(x) = \sum_{v\in [V]} p_v(x) \bm{1}_{f(x)=v}$, where $(p_v)_v$ is a realization of $(\rn{p}_v)_v$.
    \end{itemize}
\end{definition}

Our main derandomization result states that the piecewise derandomization procedure defined above yields with high probability a predictor that fools any fixed and bounded function $\phi$, in expectation over $\rn{x}\sim \D_\X$. The corresponding accuracy scales polynomially with the refinement of the partition function.

\begin{theorem}\label{thm:main-derand}
    Let $\gamma,\delta\in(0,1)$, let $\phi:\X\times [\pm1]\to[\pm1]$ and let $f$ be a $\gamma$-partition function with respect to the distribution $\D_\X$ over $\X$. Let $\rn{p}$ be any randomized predictor. Then, the $f$-piecewise derandomization of $\rn{p}$ satisfies the following with probability at least $1-\delta$:
    \[
        \Bigl| \E_{\rn{x}, \rn{p}}[\phi(\rn{x},\rn{p}(\rn{x}))] - \E_{\rn{x}}[\phi(\rn{x},{p}^f(\rn{x}))] \Bigr| \le O\bigl(\sqrt{\gamma \log(e/\delta)}\bigr)
    \]
\end{theorem}

\begin{proof}
    We view $\rn{p}^f$ as a random variable depending on the realization of $(\rn{p}_v)_v$ during the derandomization procedure. Accordingly, we also define the random variable $\rn{\Delta}$ as follows:
    \[
        \rn{\Delta} = \Bigl| \E_{\rn{x}, \rn{p}}[\phi(\rn{x},\rn{p}(\rn{x}))] - \E_{\rn{x}}[\phi(\rn{x},\rn{p}^f(\rn{x}))] \Bigr|
    \]
    We will show that with probability at least $1-\delta$ over the random choice of $(\rn{p}_v)_v$, we have $\rn{\Delta} \le O(\sqrt{\gamma\log(e/\delta)})$, as desired.
    To this end, we decompose the terms of the difference as follows:
    \begin{align*}
        \E_{\rn{x}}[\phi(\rn{x},\rn{p}^f(\rn{x}))] &= \sum_{v\in [V]}\E_{\rn{x}}[\phi(\rn{x},\rn{p}_v(\rn{x})) \bm{1}_{f(\rn{x}) = v}] =: \sum_{v\in [V]} \rn{Z}_v \\
        \E_{\rn{x}, \rn{p}} [\phi(\rn{x},\rn{p}(\rn{x}))] &= \sum_{v\in [V]}\E_{\rn{x}, \rn{p}}[\phi(\rn{x},\rn{p}(\rn{x})) \bm{1}_{f(\rn{x}) = v}] =: \sum_{v\in [V]} \mu_v
    \end{align*}
    Observe that each of the terms in the first equation is a random variable whose expectation with respect to $(\rn{p}_v)_v$ equals the corresponding term of the second equation. In particular:
    \[
        \E_{(\rn{p}_v)_v}[\rn{Z}_v] = \E_{(\rn{p}_v)_v}\E_{\rn{x}}[\phi(\rn{x},\rn{p}_v(\rn{x})) \bm{1}_{f(\rn{x}) = v}] = \E_{\rn{x}, \rn{p}}[\phi(\rn{x},\rn{p}(\rn{x})) \bm{1}_{f(\rn{x}) = v}] = \mu_v
    \]
    Moreover, we have that $|\rn{Z}_v| \le \Pr_{\rn{x}}[f(\rn{x})=v] \le \gamma$ with probability $1$ over the choice of $(\rn{p}_v)_v$, due to the definition of function $f$, and the bound on the values of $\phi$. As a consequence, we have that $|\rn{Z}_v|\le z_v$ almost surely for values $z_v$ satisfying
    \[
        \sum_{v\in [V]}z_v^2 \le  \sum_{v\in [V]}\bigl(\Pr_{\rn{x}}[f(\rn{x})=v]\bigr)^2 \le \gamma  \sum_{v\in [V]}\Pr_{\rn{x}}[f(\rn{x})=v] = \gamma
    \]
    The result follows due to a standard Hoeffding bound, by additionally observing that the random variables $(\rn{Z}_v)_v$ are independent, by the definition of $(\rn{p}_v)_v$.
\end{proof}

\paragraph{Constructing the Partition Function} We have shown that it is possible to derandomize any predictor with respect to any fixed target measure $\D_\X$ and function $\phi$, provided that there exists a $\gamma$-partition function with respect to $\D_\X$. In Appendix \ref{appendix:constructing-partition-functions}, we show that a partition function can be efficiently constructed with sample access to $\D_\X$ essentially whenever $\D_\X$ does not have point masses larger than $\gamma$. We call such distributions $\gamma$-divisible.

\begin{example}
    Any distribution over $\R^d$ that admits a density is $\gamma$-divisible for any $\gamma\in(0,1)$ and a $\gamma$-partition function can be constructed with probability at least $1-\delta$ in time $\tilde{O}(d/\gamma) \log(1/\delta)$ and using $\tilde{O}(1/\gamma)\log(1/\delta)$ samples from $\D_\X$. The same is true for any discrete distribution over $\R^d$ that does not assign mass more than $\gamma$ to any singleton.
\end{example}

\paragraph{Applications} We will now show that our main results on smooth swap multicalibration and smooth omniprediction (see \Cref{thm:main}) can be derandomized in a black-box way, as long as $\D_\X$ is $\gamma$-divisible for some appropriate $\gamma = \poly(\eps)/\log(1/\delta)$. 

\begin{remark}
    Note that the divisibility assumption is almost without loss of generality: If $\D_\X$ had heavy atoms $x_h\in \X$, then we would be able to compute accurate estimates of the Bayes-optimal values $\E[\rn{y}\,|\,\rn{x}=x_h]$ using $\poly(1/\gamma)$ samples, and apply our main algorithm on the truncation of $\D_\X$ that does not have heavy atoms. Note that this requires labeled samples (not just unlabeled) and changes the marginal distribution $\D_\X$ over features, but any  reweighting such as this one is already addressed by the very first step of our algorithm.
\end{remark}

The idea is to view the aforementioned guarantees from \Cref{thm:main} as bounds on quantities of the form: \[\sup_{\phi\in \Phi}\, \E_{\rn{x},\rn{y}}[\phi(\rn{x},\rn{y},p(\rn{x}))]\,,\]
where the function family $\Phi$ can be represented by a sparse subset $\Phi'$ of small cardinality. Therefore, it suffices to fool only the members of the family $\Phi'$, and the result follows by combining \Cref{thm:main-derand} with a union bound. 

We now define the notion of sparsification that suffices for our purposes.

\begin{definition}
    Let $\Phi$ be a collection of functions of the form $\phi:\X\times\{\pm 1\}\times [\pm 1]\to \R$. We say that $\Phi$ is $(\eps,N)$-sparsifiable with respect to a distribution $\D$ over $\X\times\{\pm 1\}$ if there is a set $\Phi'\subseteq\Phi$ with cardinality $|\Phi'|\le N$ such that for all $\phi \in \Phi$, there exists $\phi' \in \Phi'$ such that
    \[
        \sup_{p:\X\to [\pm 1]}\, \Bigl|\E_{\rn{x}, \rn{y}}[\phi'(\rn{x},\rn{y},{p}(\rn{x}))] - \E_{\rn{x}, \rn{y}}[\phi(\rn{x},\rn{y},{p}(\rn{x}))] \Bigr| \le \eps.
    \]
\end{definition}

\paragraph{Application I: Smooth Swap Multicalibration} We first focus on smooth swap multicalibration, where the class $\C$ is assumed to have bounded pseudodimension (see \cref{def:pseudodim}).

\begin{lemma}[Sparsifying Smooth Swap Multicalibration Witnesses]
\label{lem:sparsify-smooth-swap}
Let $\C\subseteq\{\X\to[\pm 1]\}$ have pseudo-dimension $d$, and let
$\Phi_{\C,L}^{\mathrm{swap}}$ be the class of functions
\[
    \phi_c(x,y,v):=c(v,x)(y-v),
\]
where $c(v,\cdot)\in\C$ for every $v\in[\pm1]$ and
$c(\cdot,x)$ is $L$-Lipschitz for every $x\in\X$ for some $L\ge 1$. Then, for every
distribution $\D$ over $\X\times\{\pm1\}$ and every $\eps\in(0,1)$,
the class $\Phi_{\C,L}^{\mathrm{swap}}$ is $(\eps,N)$-sparsifiable
with respect to $\D$, where
\[
    \log N
    =
    O\Bigl(
        \frac{dL}{\eps}
        \log\Bigl(\frac{L}{\eps}\Bigr)
    \Bigr).
\]
\end{lemma}

\begin{proof}
Let $\D_\X$ denote the $\X$-marginal of $\D$. Choose a grid
$v_1,\ldots,v_K$ in $[\pm1]$ such that every $v\in[\pm1]$ is within
$\eps/(8L)$ of some grid point, where $K=\Theta({L}/{\eps})$.
By the standard pseudo-dimension covering bound (see, e.g., \cite{haussler1992decision}), $\C$ has an
$L_1(\D_\X)$-cover of radius {$r:= \eps/8K = \Theta(\eps^2/L)$} and size at most $(1/r)^{O(d)} = (L/\eps)^{O(d)}$.

Associate with every $c$ the tuple of cover elements approximating
$c(v_1,\cdot),\ldots,c(v_K,\cdot)$, and retain one representative
$c'$ for every realized tuple. The number of retained representatives is bounded by the size of the net corresponding to each coordinate, raised to the number of coordinates $K$. Therefore, it is at most
\[
    \Bigr((1/r)^{O(d)}\Bigr)^K = \left(\frac{L}{\eps}\right)^{O(dL/\eps)}.
\]
If $c$ and $c'$ have the same tuple, then
\[
    \E_{\rn{x}}
    \left[
        \left|c(p(\rn{x}),\rn{x})-c'(p(\rn{x}),\rn{x})\right|
    \right]
    \le 2L\frac{\eps}{8L}+2Kr
    \le \frac{\eps}{2}
\]
for every predictor $p:\X\to[\pm1]$. Since
$|\rn{y}-p(\rn{x})|\le2$, it follows that
\[
\left|
\E\left[
    \phi_c(\rn{x},\rn{y},p(\rn{x}))
    -
    \phi_{c'}(\rn{x},\rn{y},p(\rn{x}))
\right]
\right|
\le\eps.
\]
Taking the supremum over $p$ proves the result.
\end{proof}

We are ready to provide a derandomization result for smooth swap multicalibration.

\begin{corollary}[Derandomizing Smooth Swap Multicalibration]
\label{cor:derandomize-smooth-swap}
Let $\C\subseteq\{\X\to[\pm1]\}$ have pseudo-dimension $d$, and suppose
that $\rn p$ is $(\C,\eps)$-smooth swap multicalibrated with respect
to $\D$. There is a universal constant $c>0$ such that, for any
$\gamma\le c\min\{1,\eps^3/(d\log(e/\eps) + \eps\log(e/\delta))\}$,
if $\D_\X$ is $\gamma$-divisible, then the piecewise derandomization
$p^f$ obtained from a $\gamma$-partition function $f$ is
$(\C,O(\eps))$-smooth swap multicalibrated with probability at least
$1-\delta$. Moreover, $f$ can be constructed using
$O((1/\gamma)\log(1/\gamma\delta))$ unlabeled samples
from $\D_\X$.
\end{corollary}

\begin{proof}
Apply Lemma~\ref{lem:sparsify-smooth-swap} with accuracy $\eps/8$ to
obtain a sparsifying set $\Phi'$ of size:
\[
    |\Phi'|\le 2^{O((d/\eps)\log(e/\eps))}
\]
    
Conditional on $f$ being a
$\gamma$-partition function, the partition-function concentration
bound (\Cref{thm:main-derand}) and a union bound over  $\Phi'$ show that the expectations of
all its functions change by at most $O(\eps)$ under derandomization.
Sparsifiability extends the guarantee to every smooth swap witness,
at an additional error of at most $\eps/4$. The guarantee for $\rn p$
therefore transfers to $p^f$. Finally, the construction of the
partition function succeeds with probability at least $1-\delta/2$;
a union bound completes the proof.
\end{proof}

\paragraph{Application II: Smooth Omniprediction} Next, we focus on smooth omniprediction, obtaining the following derandomization result.

\begin{corollary}[Derandomizing Smooth Omniprediction]
\label{cor:derandomize-smooth-omni}
Let $\rn p$ be an $(\L_{B,\beta},\C,\eps)$-omnipredictor with respect to $\D$, where $\beta>0$. There is a universal constant
$c>0$ such that, for any $\gamma\le c\min\{1,\eps^3/(\beta^3 + \beta^2\eps\log(e/\delta))\}$, if $\D_\X$ is $\gamma$-divisible, then the piecewise derandomization $p^f$ obtained from a $\gamma$-partition function $f$ is an $(\L_{B,\beta},\C,O(\eps))$-omnipredictor with probability at least $1-\delta$. Moreover, $f$ can be constructed using $O((1/\gamma) \log(1/\gamma\delta))$ unlabeled samples from $\D_\X$.
\end{corollary}

As usual, our approach will be to express the guarantee we want to preserve after derandomization as a bound on a quantity of the following form:
\[\sup_{\phi\in \Phi}\, \E_{\rn{x},\rn{y}}[\phi(\rn{x},\rn{y},p(\rn{x}))]\,,\]
where the set $\Phi$ is sparsifiable. To this end, a first idea would be to choose $\phi_\ell(x,y,v) = \ell(y,k_\ell(v))$. 
However, the class of loss functions $\L_{B,\beta}$ we consider is not guaranteed to only contain Lipschitz losses $\ell$ and it is not clear whether $\L_{B,\beta}$ is sparsifiable. 

Instead, the guarantee we have is that the optimal response function $k_\ell$ is $\beta$-Lipschitz. 
In the following lemma, we show that this suffices for our purposes, by showing that there is a choice of $\phi_\ell$ that preserves the omniprediction guarantee, and leads to a sparsifiable family $\Phi$.

\begin{lemma}[Sparsifying loss functions for smooth omniprediction]
\label{lem:sparsify-smooth-omni}
Let $\L_{B,\beta}$ be the class of naturally parametrized loss functions with a $[\pm B]$-bounded and $\beta$-Lipschitz best response function $k_\ell$. For $\ell\in\L_{B,\beta}$, define
\[
    \phi_\ell(x,y,v)
    :=
    \ell(y,k_\ell(v))-\ell(y,k_\ell(0)),
\]
and let
\[
    \Phi_{B,\beta}^{\mathrm{omni}}
    :=
    \{\phi_\ell:\ell\in\L_{B,\beta}\}.
\]
Then, for every distribution $\D$ over $\X\times\{\pm1\}$ and every
$\eps>0$, the class $\Phi_{B,\beta}^{\mathrm{omni}}$ is
$(\eps,N)$-sparsifiable with respect to $\D$, where
\[
    \log N
    =
    O\left(1+\frac{\beta}{\eps}\right).
\]
In fact, the same sparsifying set works simultaneously for every
distribution $\D$.
\end{lemma}

\begin{proof}
For $\ell\in\L_{B,\beta}$, let
\[
    r_\ell(v):=k_\ell(v)-k_\ell(0).
\]
Then $r_\ell(0)=0$ and $r_\ell$ is $\beta$-Lipschitz.

We first express $\phi_\ell$ in terms of $r_\ell$. Fix $s<t$. Since $k_\ell(t)$ minimizes $\ell(t,\cdot)$ and $k_\ell(s)$ minimizes
$\ell(s,\cdot)$, natural parametrization gives
\[
    \ell(t,k_\ell(t))
    \le
    \ell(t,k_\ell(s))
    =
    \ell(s,k_\ell(s))-(t-s)k_\ell(s),
\]
and similarly
\[
    \ell(s,k_\ell(s))
    \le
    \ell(s,k_\ell(t))
    =
    \ell(t,k_\ell(t))+(t-s)k_\ell(t).
\]
Therefore
\[
    -k_\ell(t)
    \le
    \frac{
        \ell(t,k_\ell(t))-\ell(s,k_\ell(s))
    }{t-s}
    \le
    -k_\ell(s).
\]
Since $k_\ell$ is Lipschitz, hence continuous, it follows that
\[
    \frac{d}{dv}\ell(v,k_\ell(v))
    =
    -k_\ell(v).
\]
Equivalently, recall from \cref{thm:h-vs-k} that $h_\ell(v) = \ell(v, k_\ell(v))$, viewed as a function of $v$, is concave with superderivative $h_\ell'(v) = -k_\ell(v)$, which in fact must be a derivative by $\beta$-smoothness.
Thus,
\[
    \ell(v,k_\ell(v))-\ell(0,k_\ell(0))
    =
    -\int_0^v k_\ell(t)\,dt.
\]
Using natural parametrization once more,
\[
    \ell(v,k_\ell(0))
    =
    \ell(0,k_\ell(0))-vk_\ell(0),
\]
so
\[
    \ell(v,k_\ell(v))-\ell(v,k_\ell(0))
    =
    -\int_0^v r_\ell(t)\,dt.
\]
Also,
\[
    \ell(y,a)-\ell(v,a)=(v-y)a.
\]
Hence
\[
    \phi_\ell(x,y,v)
    =
    -\int_0^v r_\ell(t)\,dt
    +(v-y)r_\ell(v).
\]
Consequently, for any $\ell,\ell'\in\L_{B,\beta}$,
\[
    \|\phi_\ell-\phi_{\ell'}\|_\infty
    \le
    3\|r_\ell-r_{\ell'}\|_\infty,
\]
since $|v|\le 1$ and $|v-y|\le 2$.

The class of $\beta$-Lipschitz functions
$r:[\pm1]\to\R$ satisfying $r(0)=0$ has an $L_\infty$-cover of
radius $\eps/6$ and size
\[
    \exp\left(
        O\left(1+\frac{\beta}{\eps}\right)
    \right).
\]
Indeed, discretize $[\pm1]$ at spacing $\Theta(\eps/\beta)$ and
quantize function values at resolution $\Theta(\eps)$. Since $r(0)=0$
and $r$ is $\beta$-Lipschitz, there are only $O(1)$ possibilities for
each successive quantized value, and there are
$O(1+\beta/\eps)$ grid points.

Partition the losses according to the closest cover element and retain
one loss from every nonempty part. If $\ell'$ is the representative
of the part containing $\ell$, then
\[
    \|r_\ell-r_{\ell'}\|_\infty
    \le
    \frac{\eps}{3},
\]
and hence
\[
    \|\phi_\ell-\phi_{\ell'}\|_\infty
    \le
    \eps.
\]
Therefore, for every predictor $p:\X\to[\pm1]$,
\[
    \left|
    \E\left[
        \phi_\ell(\rn{x},\rn{y},p(\rn{x}))
        -
        \phi_{\ell'}(\rn{x},\rn{y},p(\rn{x}))
    \right]
    \right|
    \le
    \eps.
\]
This proves the claim. Since the cover does not depend on $\D$, the
same sparsifying set works simultaneously for every distribution $\D$.
\end{proof}

We complete the proof of \Cref{cor:derandomize-smooth-omni} by using \Cref{lem:sparsify-smooth-omni}.

\begin{proof}[Proof of \Cref{cor:derandomize-smooth-omni}]
Apply Lemma~\ref{lem:sparsify-smooth-omni} with accuracy $\eps/8$ to
the centered postprocessed losses. The resulting sparsifying set has
size
\[
    \exp\left(O\left(1+\frac{\beta}{\eps}\right)\right),
\]
and each of its functions is bounded in absolute value by $O(\beta)$.
The partition-function concentration bound
(\Cref{thm:main-derand}) and a union bound therefore show that, with
probability at least $1-\delta/2$ conditional on $f$ being a
$\gamma$-partition function, every function in the sparsifying set
changes by at most $O(\eps)$ under derandomization. By
Lemma~\ref{lem:sparsify-smooth-omni}, this extends to every
$\ell\in\L_{B,\beta}$, so simultaneously for all such $\ell$,
\[
    \E\left[
        \phi_\ell(\rn{x},\rn{y},p^f(\rn{x}))
    \right]
    \le
    \E\left[
        \phi_\ell(\rn{x},\rn{y},\rn p(\rn{x}))
    \right]
    +O(\eps).
\]

Recalling that
\[
    \phi_\ell(x,y,v)
    =
    \ell(y,k_\ell(v))-\ell(y,k_\ell(0)),
\]
we obtain
\[
\begin{aligned}
    \E\left[
        \ell(\rn{y},k_\ell(p^f(\rn{x})))
        -\ell(\rn{y},k_\ell(0))
    \right]
    &\le
    \E\left[
        \ell(\rn{y},k_\ell(\rn p(\rn{x})))
        -\ell(\rn{y},k_\ell(0))
    \right]
    +O(\eps).
\end{aligned}
\]
The centering term is the same on both sides, and therefore cancels:
\[
    \E\left[
        \ell(\rn{y},k_\ell(p^f(\rn{x})))
    \right]
    \le
    \E\left[
        \ell(\rn{y},k_\ell(\rn p(\rn{x})))
    \right]
    +O(\eps).
\]
Since $\rn p$ is an $(\L_{B,\beta},\C,\eps)$-omnipredictor,
\[
    \E\left[
        \ell(\rn{y},k_\ell(\rn p(\rn{x})))
    \right]
    \le
    \min_{c\in\C}
    \E\left[
        \ell(\rn{y},c(\rn{x}))
    \right]
    +\eps.
\]
Combining the previous two inequalities gives
\[
    \E\left[
        \ell(\rn{y},k_\ell(p^f(\rn{x})))
    \right]
    \le
    \min_{c\in\C}
    \E\left[
        \ell(\rn{y},c(\rn{x}))
    \right]
    +O(\eps)
\]
for every $\ell\in\L_{B,\beta}$. Hence $p^f$ is an
$(\L_{B,\beta},\C,O(\eps))$-omnipredictor.

Finally, by the partition-function construction lemma, $f$ can be
constructed with probability at least $1-\delta/2$ using
\[
    O\left(
        \gamma^{-1}
        \left(
            \log(1/\gamma)+\log(1/\delta)
        \right)
    \right)
\]
unlabeled samples from $\D_\X$. A union bound with the derandomization
event above gives overall success probability at least $1-\delta$.
\end{proof}

\bibliographystyle{alpha}
\bibliography{refs}

\appendix

\newpage 
\section{Deferred Proofs From \texorpdfstring{\Cref{sec:upper}}{Upper Bounds Section}}
\label{sec:upper-deferred}

\begin{proof}[Proof of \Cref{thm:h-vs-k}]
    First, observe that the proper loss function $\ell(p, k_\ell(q))$ is affine in $p$ with coefficient $\partial \ell(k_\ell(q))$. Also, by definition, $h_\ell(q) = \ell(q, k_\ell(q))$. These two simple constraints imply that $\ell(p, k_\ell(q))$, viewed as a function of $p$, is an affine upper bound on $h_\ell(p)$ of the form
    \[
        \ell(p, k_\ell(q)) = h_\ell(q) + h'_\ell(q)(p-q),
    \]
    where $h_\ell' = \partial \ell \circ k_\ell$. Therefore, $h_\ell'$ is a superderivative of $h_\ell$. In this notation, it is easy to see that the divergence $D_\ell$ satisfies
    \begin{align*}
        D_\ell(p, k_\ell(q)) &= \ell(p, k_\ell(q)) - h_\ell(p) \\
        &= h_\ell(q) + h_\ell'(q)(p-q) - h_\ell(p) \\
        &= D_{h_\ell}(p, q). \qedhere
    \end{align*}
\end{proof}

\begin{proof}[Proof of \Cref{thm:general-regret}]
    The definition of $D_\ell$ states that for any action $a'$,
    \[
        \ell(\bar{p}_T, a') = \min_a\,\ell(\bar{p}_T, a) + D_\ell(\bar{p}_T, a').
    \]
    Since $\ell$ is affine in its first input and $\bar{p}_T = (w_1p_1 + \cdots + w_Tp_T)/W_T$, we may rewrite this as
    \[
        \sum_{t=1}^T w_t\, \ell(p_t, a') = \min_a \sum_{t=1}^T w_t\, \ell(p_t, a) + W_T\, D_\ell(\bar{p}_T, a').
    \]
    Separating the final term from the sum on the left gives
    \[
        \sum_{t=1}^{T-1} w_t\, \ell(p_t, a') + w_T\,\ell(p_T, a')= \min_a \sum_{t=1}^T w_t\, \ell(p_t, a) + W_T\, D_\ell(\bar{p}_T, a').
    \]
    If we choose $a' = k_\ell(\bar{p}_{T-1})$ to minimize the sum of the first $T-1$ terms on the left, this becomes
    \[
        \min_a\sum_{t=1}^{T-1} w_t\, \ell(p_t, a) + w_T\,\ell\bigl(p_T, k_\ell(\bar{p}_{T-1})\bigr)= \min_a \sum_{t=1}^T w_t\, \ell(p_t, a) + W_T\, D_\ell\bigl(\bar{p}_T, k_\ell(\bar{p}_{T-1})\bigr).
    \]
    Continuing by induction on $T$ completes the proof.
\end{proof}

Although the preceding proof of the Follow the Leader regret bound is the standard one, a helpful observation is that the proof can alternatively be phrased in terms of a potential function $W_t\, h(\bar{p}_t)$, whose change in value we track across iterations. This point of view will prove to be useful for the lower bound constructions. There, we construct examples where the algorithm requires many iterations by carefully controlling the potential gap from one iteration to the next.

\begin{proof}[Alternate Proof of \Cref{thm:general-regret}]
    Consider the superderivative $h' = \partial \ell \circ k_\ell$ of the concave function $h = h_\ell$. For any two points $u, v \in \R$, we have the first-order approximation
    \[
        h(v) - h(u) = h'(u)(v-u) - D_h(v, u).
    \]
    Substituting $u = \bar{p}_{t-1}$ and $v = \bar{p}_t$ and the update rule
    \(
        \bar{p}_t - \bar{p}_{t-1} = (w_t/W_t)(p_t - \bar{p}_{t-1})
    \)
    yields
    \[
        h(\bar{p}_t) - h(\bar{p}_{t-1}) = \frac{w_t}{W_t}h'(\bar{p}_{t-1})(p_t - \bar{p}_{t-1}) - D_h(\bar{p}_t, \bar{p}_{t-1}).
    \]
    Rearranging terms and using the fact that $\ell(p, k_\ell(q)) = h(q) + h'(q)(p-q)$ (\cref{thm:h-vs-k}) yields
    \[
        W_t\, h(\bar{p}_t) - W_{t-1}\, h(\bar{p}_{t-1}) = w_t\, \ell(p_t, k_\ell(\bar{p}_{t-1})) - W_t\, D_h(\bar{p}_t, \bar{p}_{t-1}).
    \]
    Summing over $t = 1, \ldots, T$, the left-hand side telescopes, leaving
    \[
        W_T\, h(\bar{p}_T) = \sum_{t=1}^T w_t\,\ell(p_t, k_\ell(\bar{p}_{t-1})) - \sum_{t=1}^T W_t\, D_h(\bar{p}_t, \bar{p}_{t-1}).
    \]
    Finally, observe that $W_T\, h(\bar{p}_T) = \min_a\, \sum_{t=1}^T w_t\,  \ell(p_t, a)$ by the definition of $h$.
\end{proof}

\begin{proof}[Proof of \Cref{thm:ftl-smooth}]
    By \cref{thm:h-vs-k}, the proper loss $\ell(p, k_\ell(q))$ can be written in the form
    \[
        \ell(p, k_\ell(q)) = h_\ell(q) + h'_\ell(q)(p-q),
    \]
    where $h'_\ell = \partial \ell \circ k_\ell$ is a superderivative of $h_\ell$. In the special case that $h_\ell'$ is $\beta$-Lipschitz, the function $h_\ell$ must be $\beta$-smooth, meaning that
    \[
        D_\ell(p, k_\ell(q)) = D_{h_\ell}(p, q) \le \frac{\beta}{2}(p-q)^2.
    \]
    Using the update rule
    \(
        \bar{p}_t - \bar{p}_{t-1} = (w_t/W_t)(p_t - \bar{p}_{t-1})
    \)
    and the fact that predictions are $[\pm 1]$-valued,
    \begin{align*}
        \sum_{t=1}^T W_t\, D_\ell(\bar{p}_t, k_\ell(\bar{p}_{t-1})) &\le \sum_{t=1}^T W_t \cdot \frac{\beta}{2}(\bar{p}_t-\bar{p}_{t-1})^2 \\
        &\le \sum_{t=1}^T W_t \cdot \frac{\beta}{2}\Bigl(\frac{w_t}{W_t} (p_t - \bar{p}_{t-1})\Bigr)^2 \\
        &\le 2\beta \sum_{t=1}^T \frac{w_t^2}{W_t}. \qedhere
    \end{align*}
\end{proof}

\begin{proof}[Proof of \Cref{thm:step-sizes}]
We consider the three cases separately.
\begin{itemize}
    \item \textbf{(Uniform)} When $w_t = 1$, we have
    \[
        \delta = \frac{1}{T}\sum_{t=1}^T \frac{1}{t} \le \frac{1+\log(T)}{T}.
    \]
    
    \item \textbf{(Linear)} When $w_t = t$, we have
    \[
        \delta = \frac{2}{T(T+1)}\sum_{t=1}^T \frac{2t}{t+1} \le \frac{4}{T+1}.
    \]
    
    \item \textbf{(Geometric)} When we normalize so that $W_T = 1$, we have
    \[
        w_t = \begin{cases}
            (1-\lambda)^{T-1} &\text{if }t = 1,\\
            \lambda(1-\lambda)^{T-t} &\text{if }t \ge 2,
        \end{cases}
    \]
    and hence
    \begin{align*}
        \delta &= (1-\lambda)^{T-1} + \sum_{t=2}^T \lambda^2(1-\lambda)^{T-t} \\
        &= (1-\lambda)^{T-1} + \lambda^2\Bigl(\frac{1 - (1-\lambda)^{T-1}}{1-(1-\lambda)}\Bigr) \\
        &= \lambda + (1-\lambda)^T.
    \end{align*}
    This is at most $\lambda + e^{-\lambda T}$, which is $O(\log(T)/T)$ for $\lambda = \log(T)/T$. \qedhere
\end{itemize}
\end{proof}

\begin{proof}[Proof of \Cref{thm:ftl-cal}]
    Let $W(p) = \int_0^p w(q)\,dq$. Since $w$ is $\beta$-Lipschitz, $W$ is $\beta$-smooth. Using the update rule
    \[
        \bar{p}_t - \bar{p}_{t-1} = \lambda(p_t - \bar{p}_{t-1}),
    \]
    which holds for $t \ge 2$, along with $\beta$-smoothness of $W$, we see that
    \begin{align*}
        W(\bar{p}_t) - W(\bar{p}_{t-1}) &\ge w(\bar{p}_{t-1})(\bar{p}_t - \bar{p}_{t-1}) - \frac{\beta}{2}(\bar{p}_t - \bar{p}_{t-1})^2 \\
        &\ge \lambda w(\bar{p}_{t-1})(p_t - \bar{p}_{t-1}) - \frac{\beta\lambda^2}{2}(p_t - \bar{p}_{t-1})^2.
    \end{align*}
    Summing over $t$ so that the left-hand side telescopes, and then rearranging terms, yields
    \[
        \sum_{t=2}^T w(\bar{p}_{t-1})(p_t - \bar{p}_{t-1}) \le \frac{W(\bar{p}_T) - W(\bar{p}_1)}{\lambda} + 2\beta \lambda (T-1).
    \]
    Note that the numerator on the right-hand side is bounded since $W$ is bounded. Also, the same argument goes through with $-w$ in place of $w$, completing the proof. (Note that the update rule is actually slightly different for the edge case $t = 1$, but this difference is inconsequential, contributing an additional error of at most $O(1)$ in the worst case, or $O(1/T)$ after dividing through by $T$.)
\end{proof}

\begin{proof}[Proof of \Cref{thm:loss-smoothing}]
    As in the proof of \cref{thm:ftl-smooth}, if the loss function has the form
    \[
        \ell(p, \eta) = g(\eta) - p\eta
    \]
    for some function $g$, then the proper loss $\ell(p, k_\ell(q))$ can be written as
    \[
        \ell(p, k_\ell(q)) = h(q) + h'(q)(p-q),
    \]
    where $h(q) = \inf_\eta\, \ell(q,\eta)$ is the negation of the convex conjugate of $g$ and where $h'(q) = -k_\ell(q)$ is a superderivative of $h$. Our goal is to construct slightly modified functions $g_\sigma,h_\sigma,\ell_\sigma$ that satisfy the following corresponding pair of identities:
    \begin{align*}
        \ell_\sigma(p, \eta) &= g_\sigma(\eta) - p\eta \\
        \ell_\sigma(p, k_{\ell_\sigma}(q)) &= h_\sigma(q) + h'_\sigma(q)(p-q).
    \end{align*}
    We construct these modified functions in three steps:
    \begin{enumerate}
        \item First, we set $h'_\sigma(p) = \E[h'(\rn{N}_\sigma(p))].$
        \item Next, we let $h_\sigma$ be the antiderivative of $h'_\sigma$ with $h_\sigma(0) = h(0)$.
        \item Finally, we let $g_\sigma$ be the convex conjugate of $-h_\sigma$, or more precisely
        \[
            g_\sigma(\eta) = \sup_{p \in [\pm 1]}\, p\eta + h_\sigma(p).
        \]
        This also uniquely determines $\ell_\sigma$.
    \end{enumerate}

    To build intuition for the construction of $\ell_\sigma$ from $\ell$, observe that $h'_\sigma$ is a Lipschitz and piecewise-linear approximation of $h'$. Specifically, we obtain $h'_\sigma$ from $h'$ by interpolating points on the graph of $h'$ at horizontal spacing $\sigma$. Clearly, if $h'$ is $[\pm B]$-valued, then this interpolation $h'_\sigma$ is $2B/\sigma$-Lipschitz. Accordingly, the antiderivative $h_\sigma$ is $2B/\sigma$-smooth. Moreover, one can check that $h$ and $h_\sigma$ differ pointwise by at most $B\sigma$. Indeed, by concavity, $h'$ and $h'_\sigma$ are both nonincreasing, so if we set $k = \sigma^{-1}$, then
    \begin{align*}
        |h(p) - h_\sigma(p)| &\le \sum_{j=-k}^{k-1} \int_{j\sigma}^{(j+1)\sigma} |h'(q) - h_\sigma'(q)|\,dq \\
        &\le \sum_{j=-k}^{k-1} \frac{\sigma}{2}\bigl(h'(j\sigma) - h'((j+1)\sigma)\bigr) \\
        &= \frac{\sigma}{2}(h'(-1) - h'(1)) \\
        &\le B\sigma.
    \end{align*}
    Since $h$ and $h_\sigma$ differ pointwise by at most $B\sigma$, so do their convex conjugates. In particular, $g_\sigma$ differs pointwise by at most $B\sigma$ from a convex lower bound on $g$:
    \begin{align*}
        g_\sigma(\eta) \le \sup_{p \in [\pm 1]}\, p\eta + h(p) + B\sigma \le g(\eta) + B\sigma.
    \end{align*}
    Thus, for all inputs $p \in [\pm 1]$ and $\eta \in \R$, we have
    \[
        \ell_\sigma(p, \eta) \le \ell(p, \eta) + B\sigma.
    \]
    Similarly, it is easy to check that for all inputs $p,q \in [\pm 1]$, the expected value of $\ell(p, k_\ell(\rn{N}_\sigma(q)))$ over the random noise introduced by $\rn{N}_\sigma$ is close to the smoothed proper loss $\ell_\sigma(p, k_{\ell_\sigma}(q))$:
    \begin{align*}
        \E[\ell(p, k_\ell(\rn{N}_\sigma(q)))] &= \E\bigl[h(\rn{N}_\sigma(q)) + h'(\rn{N}_\sigma(q))(p-\rn{N}_\sigma(q))\bigr] \\
        &\approx_{O(B\sigma)} h(q) + \E[h'(\rn{N}_\sigma(q))](p-q) &\text{since $h$ is $B$-Lipschitz} \\
        &\approx_{O(B\sigma)} h_\sigma(q) + h'_\sigma(q)(p-q) &\text{by definition of $h_\sigma,h'_\sigma$} \\
        &= \ell_\sigma(p, k_{\ell_\sigma}(q)). \qedhere
    \end{align*}   
\end{proof}

\begin{proof}[Proof of \Cref{thm:weight-smoothing}]
    Given $c \in \C^{\mathrm{swap}}$ and $v \in [\pm 1]$ and $x \in \X$, define
    \[
        c_\sigma(v, x) = \E[c(\rn{N}_\sigma(v), x)].
    \]
    Intuitively, $c_\sigma(v, x)$ is a piecewise-linear approximation of $c$ obtained by interpolating points on the graph of $c$ at horizontal spacing $\sigma$. Also, the function $c_\sigma$ belongs to $\C^{\mathrm{swap}}$ by our assumption throughout this paper that $\C$ is closed under convex combinations. Clearly, $c_\sigma$ is $O(1/\sigma)$-Lipschitz in its first input and
    \[
        \bigl\lvert \E[c(\rn{N}_\sigma(v), x)(y - \rn{N}_\sigma(v))] - c_\sigma(v, x)(y - v) \bigr\rvert \le O(\sigma). \qedhere
    \]
\end{proof}

\section{Deferred Proofs From \texorpdfstring{\Cref{sec:lower}}{Lower Bounds Section}}
\label{sec:lower-deferred}

\begin{proof}[Proof of \Cref{thm:geom}]
    By symmetry, $\E[\rn{x}_i^2] =  1$ and $\E[\rn{x}_i\rn{x}_j] = 0$ for $i \neq j$. Therefore,
    \begin{align*}
        \E\bigl[\langle a, \rn{x}\rangle \langle b, \rn{x} \rangle\bigr] = \E[a^\top \rn{x}\rn{x}^\top b] 
        = a^\top I_n  b 
        = \langle a, b \rangle.
    \end{align*}
    Next, we decompose $a = a^\| + a^\perp$ where $a^\| = \langle a, b \rangle b$ is parallel to $b$ and $a^\perp$ is perpendicular to $b$. By symmetry, $\langle a, \rn{x}\rangle = \langle a^\| + a^\perp, \rn{x} \rangle$ has the same distribution as $\langle a^\| - a^\perp, \rn{x} \rangle$, so
    \begin{align*}
        \E\bigl[\langle a, \rn{x} \rangle \mid \langle b, \rn{x} \rangle\bigr] &= \E\Bigl[\frac{\langle a^\| + a^\perp, \rn{x} \rangle}{2} + \frac{\langle a^\| - a^\perp, \rn{x} \rangle}{2}\mid \langle b, \rn{x} \rangle\Bigr] = \langle a^\|, \rn{x}\rangle.
    \end{align*}
    Finally, we use the standard fact that $\langle b, \rn{x} \rangle^2/n \sim \Beta(1/2, (n-1)/2)$ and hence
    \[
        C_n = \E\bigl[\lvert \langle b, \rn{x}\rangle\rvert\bigr] = \sqrt{\frac{n}{\pi}} \cdot \frac{\Gamma(n/2)}{\Gamma((n+1)/2)} \\
        \in \Bigl[\sqrt{\frac{2}{\pi}}, 1\Bigr]. \qedhere
    \]
\end{proof}

For predictors in the linear family, there is a clean characterization of valid sequences. We state it in the following lemma, which has a simple proof using \cref{thm:geom}.

\begin{lemma}
\label{thm:valid-conditions}
    Suppose that $q_1, \ldots, q_T \in \R^n$ satisfy the following conditions with respect to $p^* \in \R^n$:
    \begin{itemize}
        \item (Boundedness) $\|q_t\| \le 1$ for all $t$ and $\|q_t - q_{t-1}\| \le \lambda$ for all $t \ge 2$;
        \item (Quadratic Potential Gap)
        \(
            \frac{1}{2}\|q_t\|^2 - \frac{1}{2}\|q_{t-1}\|^2 = \lambda \, \langle q_{t-1}, \, p^* - q_{t-1} \rangle + \frac{1}{2}\|q_t - q_{t-1}\|^2.
        \)
    \end{itemize}
    Then the sequence $q_t(x) = \langle q_t, x \rangle$ is valid as per \cref{def:valid-sequence} for $\rn{x} \sim \Unif(\sqrt{n} \cdot S^{n-1})$, up to rescaling by a factor of $O(1/\sqrt{n})$.
\end{lemma}

\begin{proof}[Proof of \Cref{thm:valid-conditions}]
    Define $r_1 = q_1$ and $r_t = q_{t-1} + (q_t - q_{t-1})/\lambda$ for $t=2,\ldots, T$, which is $[\pm 1]$-valued because
    \[
        \|r_t\| \le \|q_t\| + \frac{1}{\lambda}\|q_t - q_{t-1}\| \le 1 + 1 = 2.
    \]
    Next, consider the following algebraic identity, which holds for all points $q_t, q_{t-1}$:
    \[
        \frac{1}{2}\|q_t\|^2 - \frac{1}{2}\|q_{t-1}\|^2 = \langle q_{t-1}, \, q_t - q_{t-1} \rangle + \frac{1}{2}\|q_t - q_{t-1}\|^2.
    \]
    Using the definition of $r_t$, we can rewrite this as
    \[
        \frac{1}{2}\|q_t\|^2 - \frac{1}{2}\|q_{t-1}\|^2 = \lambda\, \langle q_{t-1}, \, r_t - q_{t-1} \rangle + \frac{1}{2}\|q_t - q_{t-1}\|^2.
    \]
    Subtracting this from the potential gap condition in this lemma's statement yields
    \[
        0 = \lambda\langle q_{t-1}, \, p^* - r_t \rangle,
    \]
    which by \cref{thm:geom} is precisely the condition for a valid $q_t$ sequence from \cref{def:valid-sequence}.
\end{proof}

\begin{figure}
    \centering
    \includegraphics{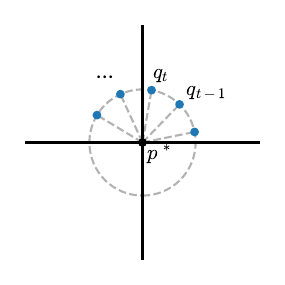}
    \includegraphics{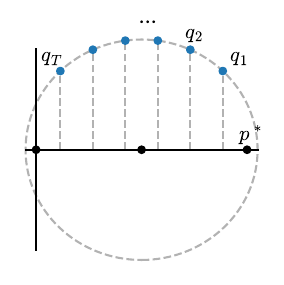}
    \includegraphics{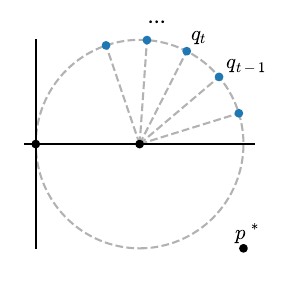}
    \caption{\cref{ex:orbit-origin,ex:arc,ex:orbit-c}.}
    \label{fig:lower-bound-examples}
\end{figure}

We now present a unified proof of our various lower bounds. We begin with two examples that give lower bounds of $\Omega(\lambda)$ and $\Omega(\min\{1, 1/(\lambda T)\})$ on degree-2 calibration error and $\Omega(\sqrt{\lambda})$ on expected calibration error. Together, they imply \cref{thm:smce-lower,thm:ece-lower}.

\begin{example}[Infinite Sequence with Small Error]
\label{ex:orbit-origin}
    Let $q_1, q_2, \ldots$ lie on a circle of radius $\rho=\sqrt{\lambda}$ centered at the origin in $\R^2$. Obtain $q_t$ by rotating $q_{t-1}$ counterclockwise about the origin by an angle $\theta$, where $\cos(\theta) = 1-\lambda$.
    By \cref{thm:valid-conditions}, this is a valid sequence because
    \begin{align*}
        \|q_t - q_{t-1}\| &= \sqrt{\|q_t\|^2 - 2\langle q_t, q_{t-1} \rangle + \|q_{t-1}\|^2} \\
        &=  \rho\sqrt{2(1-\cos(\theta))} \\
        &\le O(\lambda)
    \end{align*}
    and, letting $p^* = 0$ and $r_t = q_{t-1} + (q_t - q_{t-1})/\lambda$ as always,
    \begin{align*}
        \langle q_{t-1}, p^* - r_t \rangle &= -\Bigl(\|q_{t-1}\|^2 + \frac{\langle q_t, q_{t-1}\rangle - \|q_{t-1}\|^2}{\lambda}\Bigr) \\
        &= -\rho^2\Bigl(1 + \frac{\cos(\theta) - 1}{\lambda}\Bigr) \\
        &= 0.
    \end{align*}
    By equations \eqref{eq:twoce-formula} and \eqref{eq:ece-formula}, this sequence has degree-2 calibration error $|\TwoCE(q_t)| = \rho^2 = \Omega(\lambda)$ and expected calibration error $\ECE(q_t) = C_2\rho = \Omega(\sqrt{\lambda})$. Clearly, lower bounds on the norm $\|q_t\|$ are preserved by mixing over a random time index $\rn{t} \in [T]$, regardless of its distribution, so if we define $\rn{p}(x) = q_{\rn{t}}(x)$, it follows that $\lvert \TwoCE(\rn{p})\rvert  = \Omega(\lambda)$ and $\ECE(\rn{p}) = \Omega(\sqrt{\lambda})$, as well. Similarly, for any unbiased randomized noise function $\rn{N} : [\pm 1] \to [\pm 1]$, since $p^* = 0$, Jensen's inequality implies
    \[
        \lvert \TwoCE(\rn{N} \circ \rn{p})\rvert = \E[\rn{N}(\rn{p}(\rn{x}))^2] \ge \E\bigl[\E[\rn{N}(\rn{p}(\rn{x}))|\rn{p}(\rn{x})]^2\bigr] = \E[\rn{p}(\rn{x})^2] = \lvert \TwoCE(\rn{p})\rvert 
    \]
    and
    \[
        \ECE(\rn{N} \circ \rn{p}) = \E[\lvert \rn{N}(\rn{p}(\rn{x}))\rvert] \ge \E\bigl[\bigl\lvert \E[\rn{N}(\rn{p}(\rn{x}))|\rn{p}(\rn{x})]\bigr\rvert\bigr] = \E[\lvert\rn{p}(\rn{x})\rvert] = \ECE(\rn{p}),
    \]
    so this noise addition preserves the lower bounds on $|\TwoCE|$ and $\ECE$, as well.
\end{example}

\begin{example}[Finite Sequence with Large Error]
\label{ex:arc}
    Let $q_1, \ldots, q_T$ be a sequence of points on a circle of radius $\rho = \sqrt{1+\eps}$ centered at $p^*/2 = e_1 = (1, 0)$ in $\R^2$. By equation \eqref{eq:twoce-formula}, this circle is the set of points $q$ with $\TwoCE(q) = -\eps$. For points on this circle, $\langle q, p^*\rangle - \|q\|^2 = -\eps$, so the potential gap condition from \cref{thm:valid-conditions} becomes
    \[
        \langle e_1, q_t - q_{t-1} \rangle = -\lambda\eps + \frac{1}{2}\|q_t - q_{t-1}\|^2.
    \]
    A simple continuity argument shows that for any point $q_{t-1}$ in, say, the upper one-third of the circle, there exists a solution $q_t$ to this equation that also lies on the circle at distance $O(\lambda \eps)$ from $q_{t-1}$.\footnote{Here we require that $q_{t-1}$ lies in the upper one-third of the circle because this guarantees that a small horizontal move of $\delta = \langle e_1, q_t - q_{t-1}\rangle$ along the circle corresponds to an even smaller second-order term of $\|q_t - q_{t-1}\|^2 \le O(\delta^2)$, which is not true near the leftmost and rightmost points on the circle.} 
    Thus, for a sufficiently small universal constant $c>0$, we may choose $\eps = c \min\{1, 1/(\lambda T)\}$ and remain in the upper one-third of the circle for $T$  time steps.
    Therefore, after rescaling, this is a valid sequence whose degree-2 calibration error is negative, constant across rounds, and has absolute value $\Omega(\min\{1, 1/(\lambda T)\})$. As in the previous example, this lower bound on the absolute degree-2 calibration error is preserved even if we consider any randomized predictor $\rn{p}$ defined as a mixture of $q_1, \ldots, q_T$. Similarly, for any unbiased randomized noise function $\rn{N} : [\pm 1] \to [\pm 1]$, we have
    \begin{align*}
        \TwoCE(\rn{N} \circ \rn{p}) &= \E[\rn{N}(\rn{p}(\rn{x}))(\rn{y} - \rn{N}(\rn{p}(\rn{x})))] \\
        &= \E[\rn{p}(\rn{x})(\rn{y} - \rn{p}(\rn{x}))] - \E[\rn{N}(\rn{p}(\rn{x}))^2 - \rn{p}(\rn{x})^2] \\
        &= \TwoCE(\rn{p}) - \E[\Var(\rn{N}(\rn{p}(\rn{x}))|\rn{p}(\rn{x}))].
    \end{align*}
   Since the degree-2 calibration error was negative without noise, and since variance is always nonnegative, the above calculation shows that this noise addition preserves the lower bound on $|\TwoCE|$, as well.
\end{example} 

\begin{proof}[Proof of \Cref{thm:smce-lower}]
    The $\Omega(\lambda)$ lower bound comes from \cref{ex:orbit-origin}, and the $\Omega(\min\{1, 1/(\lambda T)\})$ lower bound comes from \cref{ex:arc}. 
    Choosing the stronger of the two examples gives that $\Omega(\min\{1, 1/(\lambda T) + \lambda\}) = \Omega(T^{-1/2})$.
\end{proof}

\begin{proof}[Proof of \Cref{thm:ece-lower}]
    The $\Omega(\sqrt{\lambda})$ lower bound comes from \cref{ex:orbit-origin}, and the $\Omega(\min\{1, 1/(\lambda T)\})$ lower bound comes from \cref{ex:arc}. 
    Choosing the stronger of the two examples gives that $\Omega(\min\{1, 1/(\lambda T) + \sqrt{\lambda}\}) = \Omega(T^{-1/3})$.
\end{proof}

Having proved \cref{thm:smce-lower,thm:ece-lower}, we now give an example of a valid predictor sequence where the degree-2 calibration error oscillates. This will prove to be a useful gadget for showing that randomization over the iterates $q_t$ is necessary.

\begin{example}[Infinite Sequence with Oscillating Error]
\label{ex:orbit-c}
    Let $q_1, q_2, \ldots$ lie on a circle of radius $\rho$ centered at $c \in \R^2$. Let $I \in \R^{2 \times 2}$ denote the identity matrix and $R_\theta \in \R^{2 \times 2}$ denote counterclockwise rotation by $\theta$. Define $q_t$ by rotating $q_{t-1}$ counterclockwise about $c$ by an angle $\theta$, and define
    \[
        p^* = \Bigl(2I + \frac{R_{-\theta} - I}{\lambda}\Bigr)c.
    \]
    As long as $\|c\| = O(1)$ and $\rho = O(1)$ and $\theta = O(\lambda)$, we have the required boundedness conditions $\|p^*\| \le O(1)$ and $\|q_t\| \le O(1)$ and $\|q_t - q_{t-1}\| \le O(\lambda)$. If we additionally set $\rho = \|c\|$ and $z_t = q_t - c$, with $r_t$ defined the same way as always, then we have conditional accuracy in expectation:
    \begin{align*}
        \langle q_{t-1}, p^* - r_t \rangle &= \Bigl\langle c + z_{t-1}, \, p^* - c - \Bigl(I + \frac{R_\theta-I}{\lambda}\Bigr)z_{t-1}\Bigr\rangle \\
        &=\bigl(\|c\|^2 - \rho^2\bigr)\Bigl(1 + \frac{\cos(\theta)-1}{\lambda}\Bigr) \\
        &= 0.
    \end{align*}
    Therefore, this is a valid sequence. Moreover, its degree-2 calibration error oscillates:
    \begin{align*}
        \TwoCE(q_t) &= \langle q_t, p^* - q_t \rangle \\
        &= \|c\|^2 - \rho^2 + \Bigl\langle q_t, \frac{R_{-\theta}-I}{\lambda}c \Bigr\rangle.
    \end{align*}
    For example, if $c = (1, 0)$ and $\rho = 1$ and $q_t = (1 + \cos(\theta t), \sin(\theta t))$ for $\theta = \lambda$ then
    \begin{align*}
        \TwoCE(q_t)
        = -\sin(\lambda t) \pm O(\lambda),
    \end{align*}
    which oscillates from $+1$ to $-1$ back to $+1$ over periods of length $O(1/\lambda)$. In this example, $p^*$ is close to $(2, -1)$ for small $\lambda$. (The predictors $p^*$ and $q_t$ are $O(1)$-bounded but not quite $[\pm 1]$-valued. Scaling down by a constant factor fixes the issue.)
\end{example}

\begin{proof}[Proof of \Cref{thm:smce-mixture}]
    For each sign $\sigma \in \{\pm 1\}$, choose  $p^*_\sigma$ and a valid sequence $q_{\sigma t}$ such that
    \[
        \langle q_{\sigma t}, \, p^*_\sigma - q_{\sigma t} \rangle = a \cdot \sin\Bigl( \sigma \frac{\pi}{4} + \lambda t\Bigr) \pm O(\lambda)
    \]
    for some constant $a > 0$. This is possible by \cref{ex:orbit-c}. Note that at any time $t$, one of the two predictors $q_{\sigma t}$ must have $\Omega(1)$ degree-2 calibration error for its respective $p^*_\sigma$, simply because $\sin(\pi/4 + \lambda t)$ and $\sin(-\pi/4+\lambda t)$ are off by a $\pi/2 = 90^\circ$ phase shift and are therefore never close to $0$ at the same time $t$. To combine these examples, define a new space $\X' = \X \times \{\pm 1\}$ with
    \begin{align*}
        p^*(x, \sigma) &= \frac{p^*_\sigma(x) + \sigma}{2} \\
        q_t(x, \sigma) &= \frac{q_{\sigma t}(x) + \sigma}{2}.
    \end{align*}
    Intuitively, the new space $\X'$ has two groups labeled $\sigma = +1$ and $\sigma = -1$, where both $p^*$ and $q_t$ take on values in the range $[0, +1]$ and $[-1, 0]$, respectively. Define the $1$-Lipschitz weight function
    \[
        w_\sigma(p) = \proj_{\bigl[\pm \frac{1}{2}\bigr]}\Bigl(p - \frac{\sigma}{2}\Bigr).
    \]
    Intuitively, the clipping to $[\pm 1/2]$ ensures that $w_{+1}$ ignores behavior in the group $\sigma = -1$ and $w_{-1}$ ignores behavior in the group $\sigma = +1$. More formally, if the group index $\sigma \in \{\pm 1\}$ is uniform, then
    \[
        \CE_{w_\sigma}(q_t) = \frac{1}{8}\langle q_{\sigma t},\, p^*_\sigma - q_{\sigma t} \rangle.
    \]
    Taking the maximum over $\sigma \in \{\pm 1\}$ gives
    \[
        \max_{\sigma \in \{\pm 1\}}\, \lvert \CE_{w_\sigma}(q_t)\rvert = \Omega(1). \qedhere
    \]
\end{proof}

\section{Deferred Proofs From Section~\ref{sec:derand}}

\begin{proof}[Proof of Theorem~\ref{thm:degree-w-cal}]
First, we prove the following helper lemma:
\begin{lemma}\label{lemma:L-lips}
    Let $w:[\pm1]\to[\pm1]$ be $L$-Lipschitz, and define $ W(v):=\int_0^v w(u)\,du.$
    Then, for every $a,b\in[\pm1]$,
    \begin{equation}\label{eq:W-lipschitz}
         \left|
            W(b)-W(a)-w(a)(b-a)
        \right|
        \leq
        \frac{L}{2}(b-a)^2.
    \end{equation}
    Moreover, for every $y\in\{\pm1\}$, the function $v\mapsto w(v)(y-v)$ is $(2L+1)$-Lipschitz.
\end{lemma}

\begin{proof}
For the first statement,
\[
\begin{aligned}
W(b)-W(a)-w(a)(b-a)
&=
\int_a^b \bigl(w(u)-w(a)\bigr)\,du.
\end{aligned}
\]
Since $w$ is $L$-Lipschitz,
\[
    |w(u)-w(a)|\leq L|u-a|,
\]
and integrating gives equation~\eqref{eq:W-lipschitz}. For the second statement, for every $u,v\in[\pm1]$,
\[
\begin{aligned}
\left|
    w(v)(y-v)-w(u)(y-u)
\right|
&\leq
|w(v)-w(u)|\,|y-v|
+
|w(u)|\,|v-u|
\\
&\leq
(2L+1)|v-u|,
\end{aligned}
\]
where we used $|y-v|\leq 2$ and $|w(u)|\leq 1$.
\end{proof}

We can now prove Theorem~\ref{thm:degree-w-cal}.
    Analogously to Theorem~\ref{thm:degree-2-cal}, we define the calibration error on the $t$-th iteration:
    \[
        \alpha_t
        =
        \mathbb{E}[w(q_t(\x))(\y-q_t(\x))].
    \]
By Lemma~\ref{lemma:L-lips} the function
$v\mapsto w(v)(y-v)$ is $(2L+1)$-Lipschitz. Since by the geometric update, 
\[
    |q_{t+1}(x)-q_t(x)|
    =
    \lambda|r_{t+1}(x)-q_t(x)|
    \leq
    2\lambda,
\]
we have
\begin{equation}\label{eq:a-t-L}
    |\alpha_{t+1}-\alpha_t|
    \leq
    (2L+1)\cdot 2\lambda
    \leq
    \alpha.
\end{equation}
Suppose toward contradiction that $|\alpha_t|>\alpha$ for every $t\in\{s,\ldots,s+T-1\}$.
Equation~\eqref{eq:a-t-L} implies that the $\alpha_t$'s cannot change sign.
Indeed, moving from a value greater than $\alpha$ to a value smaller
than $-\alpha$ would require a change greater than $2\alpha$.
Therefore, either
\begin{equation}\label{eq:a-t-case1}
    \alpha_t>\alpha
    \qquad
    \text{for every }
    t\in\{s,\ldots,s+T-1\},
\end{equation}
or
\begin{equation}\label{eq:a-t-case2}
    \alpha_t<-\alpha
    \qquad
    \text{for every }
    t\in\{s,\ldots,s+T-1\}.
\end{equation}
Here we use the following potential function to track the $q_t$ changes, which generalizes the potential function that we used in the proof of Theorem~\ref{thm:degree-2-cal}.
Namely, define
\[
    \Psi_t
    :=
    \mathbb{E}[W(q_t(x))],
    \qquad
    W(v)
    :=
    \int_0^v w(u)\,du.
\]
By Lemma~\ref{lemma:L-lips} and the geometric update rule, 
\[
    W(q_{t+1}(x))
    =
    W(q_t(x))
    +
    \lambda w(q_t(x))(r_{t+1}(x)-q_t(x))
    +
    e_t(x),
\]
where $|e_t(x)|\leq 2L\lambda^2$ pointwise.

By the query-response pair property $\mathbb{E}[\y-r_{t+1}(\x)\mid q_t(\x)] = 0$, it follows that for every bounded function $w: [\pm 1] \rightarrow \mathbb{R}$, 
\begin{align*}
    \E[w(q_t(\x))(\y-r_{t+1}(\x))] &= \E[\E[w(q_t(\x))(\y-r_{t+1}(\x)) \mid q_t(\x)]] \\
    &= \E[w(q_t(\x))\E[\y-r_{t+1}(\x) \mid q_t(\x)]] \\
    &= 0.
\end{align*}
Hence,
    \begin{align*}
        \alpha_t
        &=
        \mathbb{E}[w(q_t(\x))(\y-q_t(\x))]
        \\
        &=
        \mathbb{E}[w(q_t(\x))(\y-r_{t+1}(\x))]
        +
        \mathbb{E}[w(q_t(\x))(r_{t+1}(\x)-q_t(\x))]
        \\
        &=
        \mathbb{E}[w(q_t(\x))(r_{t+1}(\x)-q_t(\x))].
    \end{align*}
Then, taking expectations, and by the definition of the potential function,
\begin{align*}
        \Psi_{t+1}-\Psi_t
        &=
        \lambda
        \mathbb{E}[w(q_t(\x))(r_{t+1}(\x)-q_t(\x))]
        +
        \mathbb{E}[e_t(\x)]
        \\
        &=
        \lambda \alpha_t+\mathbb{E}[e_t(\x)].
    \end{align*}
    Moreover, $|\mathbb{E}[e_t(\x)]|\leq \mathbb{E}[|e_t(\x)|]\leq 2L\lambda^2.$
    Therefore,
    \begin{equation}
        \lambda \alpha_t-2L\lambda^2
        \leq
        \Psi_{t+1}-\Psi_t
        \leq
        \lambda \alpha_t+2L\lambda^2.
        \label{eq:w-cal-potential-increment}
    \end{equation}

As in the case of degree-2 calibration, summing the potential identity over consecutive rounds makes the intermediate potential values cancel:
\[
    \sum_{t=s}^{s+T-2}(\Psi_{t+1}-\Psi_t)
    =
    \Psi_{s+T-1}-\Psi_s.
\]
Consequently, rearranging gives
\[
    \frac{1}{T-1}\sum_{t=s}^{s+T-2}\alpha_t
    =
    \frac{\Psi_{s+T-1}-\Psi_s}{\lambda(T-1)}
    -
    \frac{1}{\lambda(T-1)}
    \sum_{t=s}^{s+T-2}\mathbb{E}[e_t(x)].
\]
Since the weight function is bounded in absolute value by one, its integral is 1-Lipschitz.
Together with the bounded prediction range and the remainder estimate above, this gives
\[
    |\Psi_{s+T-1}-\Psi_s|\leq 2,
    \qquad
    |\mathbb{E}[e_t(x)]|\leq 2L\lambda^2.
\]
Taking absolute values therefore yields
\[
    \begin{aligned}
        \left|
        \frac{1}{T-1}\sum_{t=s}^{s+T-2}\alpha_t
        \right|
        &\leq
        \frac{|\Psi_{s+T-1}-\Psi_s|}{\lambda(T-1)}
        +
        \frac{1}{\lambda(T-1)}
        \sum_{t=s}^{s+T-2}|\mathbb{E}[e_t(x)]| \\
        &\leq
        \frac{2}{\lambda(T-1)}+2L\lambda,
    \end{aligned}
\]
which is the bound used in the proof sketch in Section~\ref{sec:L-lipschitz-cal}.
    
By assumption, $\lambda\leq\alpha/(4L+2),$ and hence $2L\lambda^2 \le \lambda \alpha/2.$
Suppose first that equation~\eqref{eq:a-t-case1} holds.
From equation~\eqref{eq:w-cal-potential-increment} it follows that 
\begin{align*}
    \Psi_{t+1}-\Psi_t
    &\geq
    \lambda \alpha_t-2L\lambda^2
    \\
    &>
    \lambda\alpha-2L\lambda^2
    \\
    &\geq
    \frac{\lambda \alpha}{2}.
\end{align*}
Hence, summing over $t=s,\ldots,s+T-2$ gives $\Psi_{s+T-1}-\Psi_s  >(T-1)(\lambda\alpha/2) > 2.$
This is impossible: since $|w|\leq1$, the function $W$ is $1$-Lipschitz on $[\pm 1]$, and so for every $u,v\in[\pm 1]$, $|W(v)-W(u)| \leq |u-v| \leq 2$. 
Hence,
\begin{align*}
    |\Psi_{s+T-1}-\Psi_s|
    &=
    \left|
        \mathbb{E}\left[
            W(q_{s+T-1}(\x))-W(q_s(\x))
        \right]
    \right|
    \\
    &\leq
    \mathbb{E}\left[
        \left|
            W(q_{s+T-1}(\x))-W(q_s(\x))
        \right|
    \right]
    \\
    &\leq
    2,
\end{align*}
and hence it cannot be that $\Psi_{s+T-1}-\Psi_s > 2$.

Symmetrically, if we instead suppose that equation~\eqref{eq:a-t-case2} holds, from equation~\eqref{eq:w-cal-potential-increment} we get that
\[
    \Psi_{t+1}-\Psi_t \leq -\frac{\lambda\alpha}{2}.
\]
Summing over $t=s,\ldots,s+T-2$ gives $\Psi_{s+T-1}-\Psi_s<-(T-1)\frac{\lambda\alpha}{2}<  -2.$
This again contradicts the fact that $|\Psi_{s+T-1}-\Psi_s|\leq2$.
Hence the entire block cannot satisfy $|\alpha_t|>\alpha$.

Therefore,
there exists some $t\in\{s,\ldots,s+T-1\}$ such that
\[
    \left|
        \mathbb{E}[w(q_t(\x))(\y-q_t(\x))]
    \right|
    \leq
    \alpha,
\]
as we wanted to show.
\end{proof}

\section{Constructing a Partition Function}\label{appendix:constructing-partition-functions}

We now argue that partition functions (\Cref{def:partition-function}) exist and can be constructed efficiently from unlabeled samples from $\D_\X$ under the following mild conditions.

\begin{definition}\label{definition:divisible-distribution}
    We say that a distribution $\D_\X$ is $\gamma$-divisible if:
    \begin{itemize}
        \item \textbf{(No Heavy Atoms)} For all $x\in \X$, $\Pr_{\rn{x}}[\rn{x}= x] \le \gamma$.
        \item \textbf{(Total Order)} There is a total order $\prec$ on $\X$ admitting an efficient comparison oracle.\footnote{We also assume the standard measurability conditions required for the VC $\eps$-net theorem. These mild measurability conditions hold for all distributions on finite or countable domains, and for all distributions on $\R^d$ admitting a density, using the lexicographic order.}
    \end{itemize}
\end{definition}

Essentially, the only non-trivial requirement we impose on the distribution $\D_\X$ is that there are no points with probability mass larger than $\gamma$. The existence of a total order is, for example, always true for distributions over $\R^d$ that admit a density or a probability mass function. Under these conditions, we show that a $\gamma$-partition function can be constructed as follows.

\begin{lemma}
    Suppose that $\D_\X$ is $\gamma$-divisible. Then, there is an algorithm that, upon receiving $m=O((1/\gamma)\log(1/\gamma\delta))$ i.i.d. samples from $\D_\X$, outputs with probability at least $1-\delta$ a $\gamma$-partition function. The algorithm runs in time $O(mt\log(m))$, where $t$ is the time required for a single evaluation of the comparison oracle on $\X$.
\end{lemma}

\begin{proof}
    Let $S$ be the set of $m$ i.i.d. samples from $\D_\X$. Let $\cal{R}$ be the set of singletons and intervals with respect to the order $\prec$, i.e., sets of the form:
    \[
        \{x\in\X: x_-\prec x \prec x_+\} \quad \text{or} \quad \{x_0\}, \quad\text{ where }\quad x_-,x_+\in\X\cup\{-\infty,\infty\},\quad x_0\in \X
    \]
    Observe that the VC dimension of $\cal{R}$ is $2$. Therefore, due to the standard VC $\gamma$-net theorem (see, e.g., \cite{kupavskii2020epsilon} and references therein), we have that $S$ is a $\gamma$-net for $(\X,\cal{R})$ with probability at least $1-\delta$.

    Let $x_1 \prec x_2 \prec \dots \prec x_m$ be a sorting of the elements of $S$, which can be computed in time $O(mt\log m)$. Without loss of generality, we assume that there are no duplicates, since we can otherwise remove them. We define $f$ as follows:
    \[
        f(x) = 
        \begin{cases}
            1, \text{ if }x\prec x_1 \\
            i, \text{ if }x_{i-1}\prec x \prec x_i \\
            m+1, \text{ if }x_m \prec x \\
            m+i+1, \text{ if }x=x_i
        \end{cases}
    \]
    Observe that $f$ is measurable and total. Moreover, for any $i\in[m]$, $f$ takes the value $m+i+1$ with probability at most $\gamma$, because $\D_\X$ does not have heavy atoms.

    To complete the argument, note that $S$ does not contain any points in the intervals of the form $(-\infty,x_1), (x_{i-1},x_i),(x_m,\infty)$. Since $S$ is a $\gamma$-net (w.h.p.), this implies that the mass of $\D_\X$ on each of these intervals is at most $\gamma$.
\end{proof}

\begin{example}
    Any distribution over $\R^d$ that admits a density is $\gamma$-divisible for any $\gamma\in(0,1)$. We may use the lexicographic order, whose comparisons can be performed in time $O(d)$.
\end{example}

\section{Implementation using Statistical Queries}\label{appendix:sq-preservation}

In this section, we demonstrate that all of our main results in \Cref{thm:main-al,thm:main} continue to hold in the statistical query framework of \cite{kearns98sq}. The statistical query framework is a restricted computation model for statistical problems. It captures a wide range of algorithms like moment methods, convex relaxations, polynomial regression, and gradient descent \cite{feldman2017statistical,FeldmanGV17}. Concretely, it restricts the access of the algorithm to the underlying distribution $\D$ as follows.

\begin{definition}[Statistical Query Algorithm, \cite{kearns98sq}]\label{definition:sq}
For a labeled distribution $\D$ over $\X\times\{\pm 1\}$ and tolerance $\tau>0$, we say that an algorithm is a statistical query algorithm with tolerance $\tau$ if it has access to $\D$ only through adaptive statistical queries $\phi:\X\times\{\pm 1\}\to[-1,1]$ of tolerance $\tau$. Concretely, in response to $\phi$, the algorithm receives a value $v$ satisfying
\[
    \bigr|v-\E[\phi(\x,\y)]\bigr|\le \tau.
\]
\end{definition}

In order to show that our reductions are SQ-preserving, we observe that the only part where we access the distribution $\D$ is when we invoke the conditional multiaccuracy oracle of \Cref{thm:cond}. It therefore suffices to prove the following lemma.

\begin{lemma}\label{lemma:sq-preservation}
    Suppose that we have a $(\C,\eps)$-multiaccuracy oracle that can be implemented with $\tau$-tolerant statistical queries. Then, we have an algorithm for conditional multiaccuracy with the specifications of \Cref{thm:cond}, that can be implemented with $\tau$-tolerant statistical queries.
\end{lemma}

\begin{proof}
    Suppose that we have access to statistical queries of tolerance $\tau$ under a labeled distribution $\D$ over pairs $(\x,\y)$. The conditional multiaccuracy oracle of \Cref{thm:cond} works by calling \Cref{alg:reweight} a number of times using the same distribution $\D$, but different label reweightings $m$. For each reweighting, the multiaccuracy oracle is called on distributions of the form $(\x,\z)$, where:
    \[
        \E[\z|x]=m(x)\cdot \E[\y|x]+(1-m(x))\cdot p(x)\,,
    \]
    for some known $p:\X\to [-1,1]$.

    Therefore, to implement a $\tau$-tolerant statistical query $\phi$ requested by the multiaccuracy oracle, we observe:
    \begin{align*}
        \E[\phi(\x,\z)] &= \E\Bigr[\frac{1+\z}{2}\cdot \phi(\x,1)+\frac{1-\z}{2}\cdot \phi(\x,-1) \Bigr] \\
        &= \frac{1}{2}\E\Bigr[\phi(\x,1)+\phi(\x,-1)+ \Bigr(m(\x)\cdot \y+(1-m(\x)) \cdot p(\x)\Bigr)\cdot\Bigr(\phi(\x,1)-\phi(\x,-1)\Bigr) \Bigr]
    \end{align*}
    Note that the quantity inside the expectation is a function of $(\x,\y)$, and is pointwise bounded by $2$. Therefore, if we define $\phi'$ as follows:
    \[
        \phi'(\x,\y)= \frac{1}{2}\cdot \Bigr(\phi(\x,1)+\phi(\x,-1)+ \Bigr(m(\x)\cdot \y+(1-m(\x)) \cdot p(\x)\Bigr)\cdot\Bigr(\phi(\x,1)-\phi(\x,-1)\Bigr)  \Bigr)\,,
    \]
    then we have $|\phi'(\x,\y)|\le 1$, and we can obtain a $\tau$-tolerant query for $\phi'$ using the available oracle. Let $v'\in[-1,1]$ be the response of the oracle. Then:
    \begin{align*}
        \Bigr| v' - \E[\phi(\x,\z)] \Bigr| = \Bigr| v' - \E[\phi'(\x,\y)] \Bigr|  \le \tau
    \end{align*}
    Therefore, we can implement a statistical query for $(\x,\z)$ using a statistical query for $(\x,\y)$ with the same tolerance.
\end{proof}

As an immediate consequence of \Cref{lemma:sq-preservation}, we obtain that our main results in \Cref{thm:main-al,thm:main} are SQ-preserving.

\end{document}